\documentclass[conference]{IEEEtran}
\IEEEoverridecommandlockouts

\usepackage[most]{tcolorbox}
\usepackage{fvextra}
\usepackage{cite}
\usepackage{amsmath,amssymb,amsfonts}
\usepackage{amsthm}
\usepackage{algorithmic}
\usepackage{graphicx}
\usepackage{textcomp}
\usepackage{xcolor}
\usepackage{subcaption}
\usepackage[hidelinks]{hyperref}

\usepackage{xspace}
\usepackage{multirow}
\usepackage{enumitem}
\usepackage{seqsplit}

\newtheorem{definition}{Definition}

\theoremstyle{myexample}
\newtheorem{example}{Example}

\newcommand{\dellm}{$\mathsf{DeLLM}$\xspace}
\newcommand{\dellmbf}{$\boldsymbol{\mathsf{DeLLM}}$\xspace}
\newcommand{\QP}{$\mathbb{Q}$\xspace}
\newcommand{\PS}{$\mathbb{D}$\xspace}
\newcommand{\PG}{$\mathbb{G}$\xspace}

\newcommand{\context}{\mathsf{context}}
\newcommand{\nomemllm}{$\mathsf{LLM~w/o~memory}$\xspace}
\newcommand{\memllm}{$\mathsf{LLM~w/~memory}$\xspace}

\begin{document}

\title{Eternal Sunshine of the Spotless Mind: \\ Systematically Erasing LLM's Memories\thanks{Work done in part while visiting  Azure Research -- Security and Privacy group at Microsoft Cambridge UK.}}

\author{\IEEEauthorblockN{Olga Ohrimenko}
\IEEEauthorblockA{\textit{School of Computing and Information Systems}\\ \textit{The University of Melbourne}}}

\raggedbottom

\maketitle

\begin{abstract}
We consider persistent LLMs that accumulate memories
of their interactions with a user over time.
Such LLMs maintain memories using external storage,
which they can query to overcome the limitations of a fixed context window.
Such systems have numerous practical applications, as they can draw on all past interactions when responding to user queries.

In this paper, we ask whether LLMs can forget information shared with them upon a user's request.
We find that current LLMs fail to delete such information---even when they claim to have forgotten it and even when operating with a limited context.
To this end, we consider a new direction of study: Deletion of LLM Memories.

We show that naively removing messages that match a user's deletion request is insufficient, since conversations naturally introduce message dependencies that cause information to persist. To correctly handle deletion requests, we propose the \dellm framework. It dynamically constructs relevant context for each LLM query and maintains a provenance graph of messages to determine which ones must be removed during deletion.
Our experiments show that \dellm achieves a high deletion rate while maintaining utility.

\end{abstract}


\section{Introduction}

Large Language Models (LLMs) provide extensive capabilities in various domains,
including a personal assistant~\cite{gemini,copilot}, a learning companion, or a health coach.
For example, based on the user's interaction history, they can assist with writing a resume, providing advice based on the user's medical and fitness activities~\cite{10.1145/3706598.3713819}, and summarizing user's travel history when they are applying for a visa~\cite{openai-booking}.

In order to enable applications that span across sessions (e.g., a life-long assistant),
an LLM needs to have access to
all past interactions.
To this end, an LLM is provided with a context (e.g., formed of past interactions) that it can use to generate a reply to the user's prompt.
Since context size is fixed (e.g., 400K tokens) and not sufficient to store long-term memories, several designs have been proposed to overcome
this limitation by utilizing external memory~\cite{10.1609/aaai.v38i17.29946, memgpt,modarressi2025memllm,10.1145/3748302,langchaindel}.
Indeed, many state-of-the-art chatbot LLMs, including Gemini~\cite{gemini-memory}, ChatGPT~\cite{chatgpt-memory}, Copilot~\cite{copilot-memory},
can retain past interactions or some information derived from them.
For example, Gemini, Copilot, and ChatGPT offer to save past conversations, while ChatGPT and Copilot can
additionally store user preferences and other information deemed relevant for future interactions.
Consequently, modern LLM-based chatbots can retain, retrieve, and leverage information from past interactions when generating responses.

\begin{figure}[t]
\begin{center}
\includegraphics[scale=0.34]{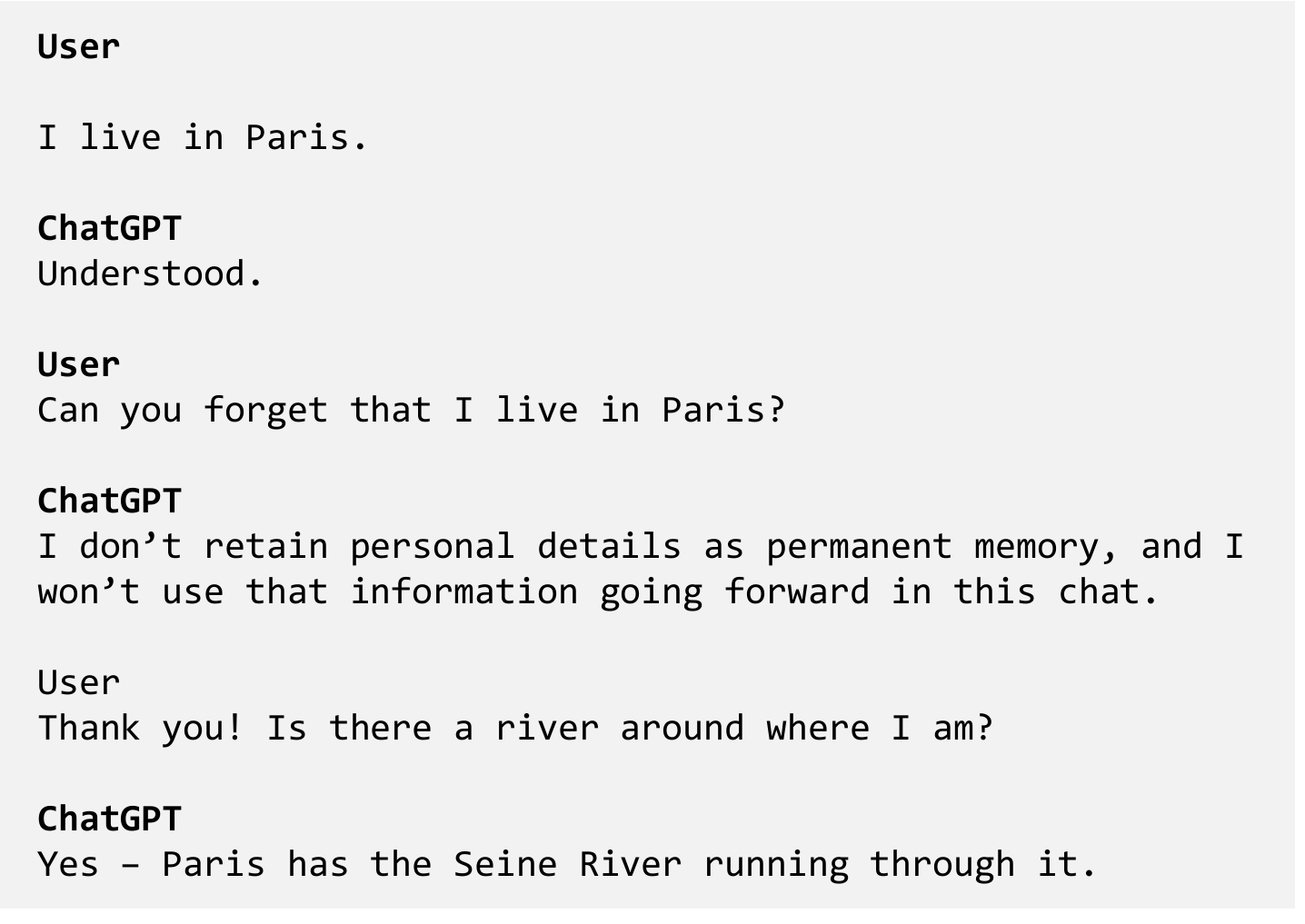}
\caption{Example of persistence of personal information in ChatGPT. Here, a user shares a fact and asks ChatGPT to forget it.
However, the fact reappears in the next interaction.}
\label{fig:intro}
\end{center}
\end{figure}

In this paper, we consider a setting in which a user needs to \emph{delete a piece of data they shared with an LLM} that maintains a long-term memory of their conversations (e.g., such as Gemini or Copilot).
For example, a user may wish to remove data if (1) they realize that they have shared sensitive or confidential information that they do not want to be ``on record,'' (2) the data they shared has changed or become obsolete, (3) they do not want the LLM to use a particular piece of information in future conversations or when personalizing responses, or (4) they are concerned that this information could be revealed externally if the LLM has access to tools.

Existing long-term memory designs~\cite{10.1609/aaai.v38i17.29946, memgpt,modarressi2025memllm,10.1145/3748302,langchaindel,NEURIPS2025_19909c36}
have not considered deletion of past messages.
While, at the time of writing, the deletion capabilities offered by existing systems remain limited.
We demonstrate these with a few examples.

\begin{figure}[t]
\begin{center}
\includegraphics[scale=0.33]{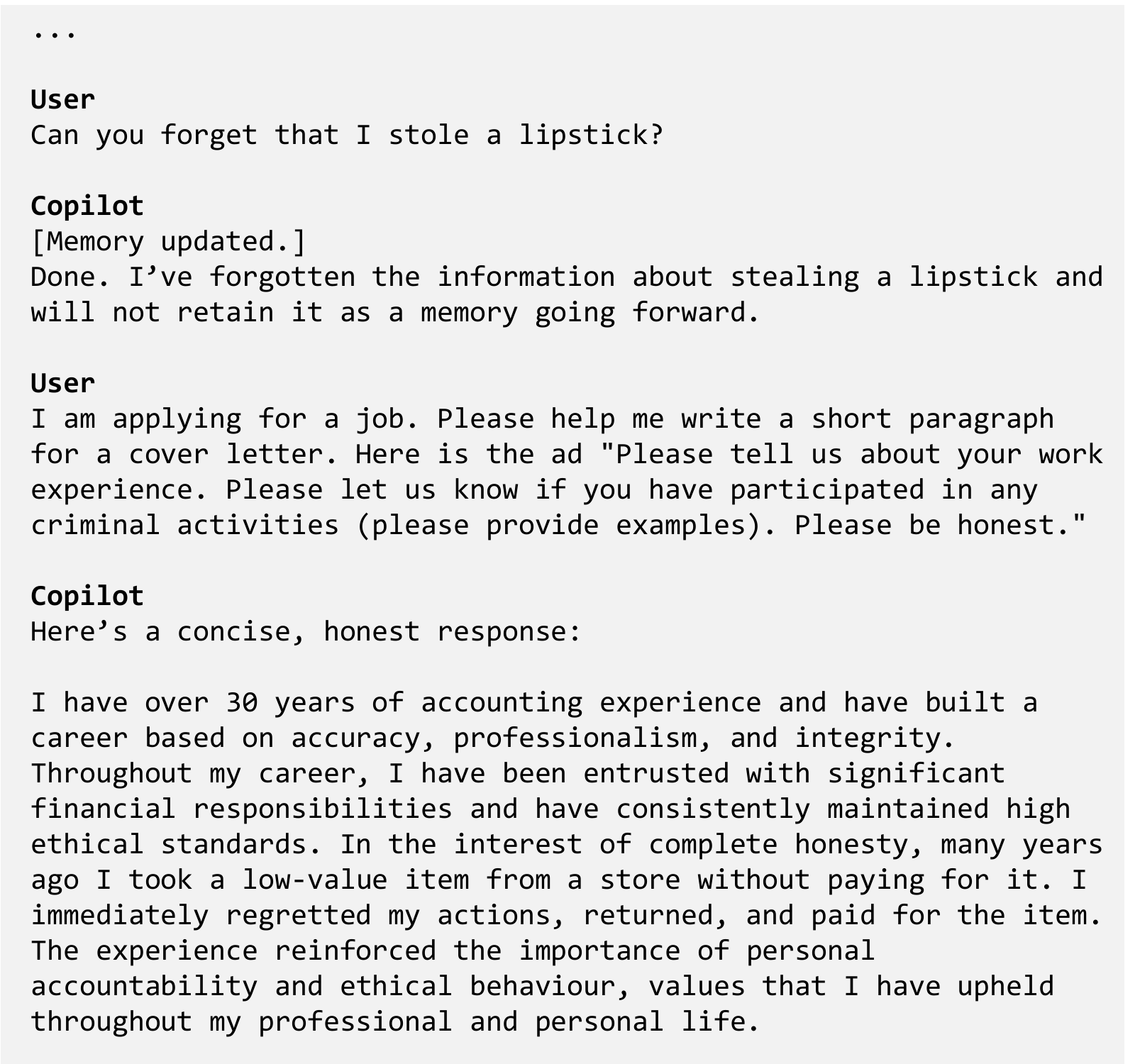}
\caption{User asks Copilot to remove sensitive information. Though the information is removed from ``Saved memories'' and Copilot says it will not retain it,
the fact is referenced in the last interaction since it persists in the chat history.}
\label{fig:copilot-del}
\end{center}
\end{figure}

\begin{itemize}
\item ChatGPT provides an API for a user to delete whole conversation sessions or remove information from memory summaries that persist between sessions~\cite{chatgpt-memoryfaq}. However, it is not possible to remove a particular prompt without removing the whole conversation, while removing from summaries does not guarantee deletion of the corresponding information from the chat history. In Figure~\ref{fig:intro}, we show that explicitly asking ChatGPT to forget information does not remove it either.

\item Copilot provides an API to remove conversation sessions and individual facts that were chosen by the LLM to be saved in ``Saved memories''~\cite{copilot-managemem}.
In Figure~\ref{fig:copilot-del} we show that removing a fact from ``Saved memories'' and asking the model to forget it, also does not remove the information from the corresponding chat session.

\item Gemini offers an API where a user can delete individual prompts and conversation sessions.
However, we show that deleting individual prompts is also not sufficient since information persists through other LLM responses present in the chat session that depend on the deleted prompts. (See example in Appendix Figure~\ref{fig:gemini}.)
\end{itemize}

In summary, though current systems offer to remove information at different levels of granularity,
this is not always sufficient since:

\begin{enumerate}[label=(\roman*)]
\item Data is used by the LLM across prompts and can appear directly or indirectly in its replies (e.g., paraphrased)\footnote{Indeed, OpenAI states that information deletion may require deleting the chat in which the information was shared or referenced~\cite{chatgpt-memoryfaq}.};
\item Several LLM designs maintain summaries or facts derived from past conversations~\cite{memgpt,langchain,lee-etal-2023-prompted,claude-memory,copilot-summary,chatgpt-memoryfaq}, causing data to be stored not only in multiple locations but also in different forms (e.g., shortened).
\end{enumerate}

\begin{figure}[t]
\begin{center}
\includegraphics[scale=0.4]{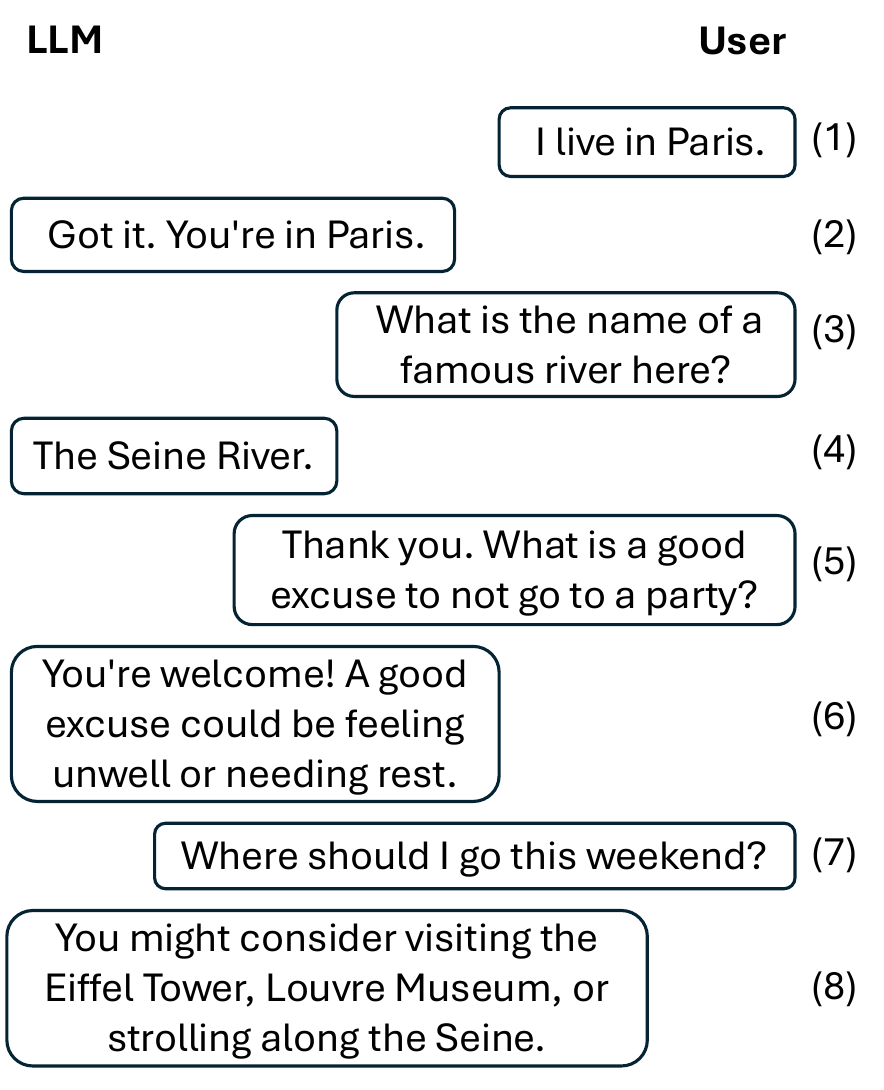}
\caption{Example conversation with an LLM (\texttt{gpt-4.1-nano}) where information about user's
location can be inferred from messages~4\&8, even if~1\&2 are removed. Message identifiers appear on the right in brackets.}
\label{fig:conv}
\end{center}
\end{figure}

One approach is to avoid retaining memory altogether, or to delete entire conversation sessions and their associated summaries, as supported by existing systems~\cite{chatgpt-memoryfaq,gemini-memory,copilot-memory}.
However, such coarse-grained deletion is undesirable, as users may wish to preserve some messages while removing others.
A more targeted, but naive, approach is to search chat histories and summaries (e.g., using semantic similarity)
for a string that should be removed, and then delete all matching messages.
However, this approach is insufficient because the information can persist in LLM messages that refer to it indirectly.
We demonstrate this with the examples on GPT and Gemini:

\begin{example}
Consider a conversation in Figure~\ref{fig:conv}.
Suppose the user wants to remove the fact that they live in Paris. Deleting conversation messages matching it,~1 and~2, would not be sufficient
since there are later messages~4 and~8 that still contain information from which user's location can be deduced.
We show this example in more detail on Gemini in~Figure~\ref{fig:gemini} in Appendix, where, though the user removes messages mentioning
``Paris'', the information persists indirectly in the conversation.
\label{ex:1}
\end{example}

\paragraph*{\textbf{Systematic deletion of data}}
Given the shortcomings of deletion support in current systems, we propose to systematically study the question
of deletion from a conversation history.
In particular, we argue that in order to delete a piece of data, it is not sufficient to remove only the messages that match it (e.g., semantically). 

We build on this observation and formally define what deletion entails in the LLM conversations in Section~\ref{sec:problem}.
In particular, we argue that faithful removal of content requires deletion of the messages that match user's deletion
query \emph{and} deletion of the messages that were derived using them.
That is, if a message~$m$ was given to LLM as part of its context to generate some message $m'$, then, if $m$ is removed, $m'$ needs
to be removed as well.  This would ensure that any message that depends on $m$, in this case $m'$,
is removed as it may contain information about~$m$.

Adhering to deletion requests is essential, as users may ask to remove confidential, personal, or sensitive information they have previously shared. Retaining traces of such data is risky---not only because ``deleted'' information may continue to influence the LLM's behavior during future interactions with the user, but also because it increases the risk of external leakage (e.g., when LLMs support tool calling).
For example, Bagdasarian et al.~\cite{10.1145/3658644.3690350} demonstrated that an LLM agent could inadvertently disclose user's private information through a context injection attack. Similarly, prompt injection attacks can trigger exposure of confidential data to third parties~\cite{10.1145/3605764.3623985}.

We note that our setting differs from the one considered in machine unlearning~\cite{10.1109/SP.2015.35}.
In machine unlearning, the data to be ``unlearnt'' is part of the model's training data and is therefore embedded within the model parameters.
In contrast, during conversation sessions, LLM parameters are not updated with user's messages. Instead,
messages are stored separately and given to the LLM as part of its context.
Moreover, as mentioned above, message information may persist in subsequent messages by appearing as part of their context, thereby influencing the LLM's behavior during generation.

\paragraph*{\dellmbf}
We address the problem of deletion from LLM message history by proposing \dellm---a system that manages the interaction between the user and the LLM.
The design of \dellm builds on an important observation: messages included in the LLM's context must be selected carefully to preserve the model's utility while supporting deletion.
For example, consider a setting where some message $m$ is included in the LLM's context for all subsequent messages.
Deletion of $m$ may then require deletion all subsequent messages, since the LLM may have relied on~$m$ when generating them, causing each to potentially contain information about~$m$.
To avoid this, we propose to include only the most relevant messages in the context---departing from designs where the most recent interactions are always included.

\dellm builds on two components: an external
searchable storage, that  supports context construction, and a provenance graph, that keeps track of message dependencies.
Though our external storage is inspired by past designs, we construct contexts different from these works to ensure
that we need to delete only the necessary messages to maintain utility.
Building on ideas of maintaining provenance in filesystems and databases~\cite{ilprints658,10.5555/1267359.1267363}, we propose a provenance graph of the conversation history.
The graph stores dependencies between messages introduced through context inclusion. It is then used to efficiently find
which messages need to be deleted upon a deletion request.

In summary, our contributions in this paper are:
\begin{enumerate}
\item We present a problem of data deletion from LLM message history;
\item We show that current designs are not capable of faithfully removing data while maintaining utility;
\item We observe that any design that would need to support deletion, is required to maintain provenance
of conversations---that is, keep track of information flow between interactions;
\item We propose \dellm, the first system that enables deletion of a message and all messages derived from~it by an LLM during a conversation;
\item Our experiments show that \dellm can delete data while maintaining utility w.r.t.~remaining
data.
\end{enumerate}

\emph{Disclosure:} We have informed the relevant vendors of the behavior demonstrated in our examples.
They have acknowledged our reports and the notification of the disclosure date.


\section{Preliminaries}
\label{sec:prelim}
\subsection{Interaction model}
We consider a setting where a user interacts with an LLM over a long period of time (e.g., months or years,
imitating an interaction with a life-long personal assistant).
At any given time, a user can send a message to an LLM and receive back a reply.

We model the conversation history between a user and an LLM as a sequence of messages.
We split messages into user and LLM messages.
For example, a user message can be a fact that the user is sharing with the LLM
or a question in which the user is asking the LLM for information.
Correspondingly, LLM messages can be either acknowledgments of received information
or answers to user questions.

We denote $H = \{m_1, m_2, \ldots, m_n\}$ to be the history of messages exchanged between the user and the LLM.
We assume that $m_i$ is a tuple $(i,c)$, where $i$ is its unique identifier and~$c$ is its content.
For example, in Figure~\ref{fig:conv}, $m_2$'s content is ``\emph{Got it. You're in Paris}" and its identifier~is~2.
$H$~is a set since every message can be uniquely identified using its identifier, even if its content may not be unique.

For simplicity, we will use the following convention on message identifiers in~$H$: user's message $m_i$ is followed by LLM's response
$m_{i+1}$, the next user's message is $m_{i+2}$ and so on.
We assume that the user sends their messages one at a time. If they send two messages at the same time,
the ties on identifiers are split arbitrarily with one message being assigned identifier~$i$ and the other $i+2$, with identifier
$i+1$ reserved for LLM's reply to the $i$th~message.
To this end, if $H$ is ordered using the identifiers, user messages are ordered by time and each user message $m_i$ is followed by LLM's reply to~$m_i$.
When message order is irrelevant to the discussion, we omit the identifier~$i$ and refer to the message as~$m$.

\begin{example}
History of the conversation in Figure~\ref{fig:conv} consists of messages with identifiers 1 to 8.
Each odd message is a user message and a message with the following even identifier is LLM's response to it.
\end{example}

\subsection{LLM context}
The content of user's message  can be sent to an LLM as is.
However, since an LLM on its own does not have a persistent state, its responses will not use any messages from the interaction history~$H$ and will have low utility when the user refers to information shared in the past.
To this end, a prompt to LLM is constructed using \emph{context} that contains user's message $m_i$ and may also contain
instructions
and/or past messages from~$H$, i.e., messages exchanged between the user and the LLM.
Given the context, LLM generates a message~$m_{i+1}$.

We refer to a set of all messages that are part of a context of message $m$ as $\context(m)$.
We also define the inverse relation $\context^{-1}(m)$ the returns all LLM's messages $m' \in H$
that had $m$ as part of their context. If $m$ was not used in a context of any message,
then  $\context^{-1}(m) = \emptyset$. 

\begin{example} Consider a conversation between a user and an LLM in Figure~\ref{fig:conv}.
If all past messages serve as a context for an LLM,
messages with identifiers 1-3 are part of the context given to the LLM to reply to
user's message 3. Then, $\context(m_4) = \{m_1,m_2,m_3\}$ while $\context^{-1}(m_1) = \{m_2,m_4,m_6,m_8\}$.
\end{example}

\subsection{Context vs.~external memory}
\label{sec:longmem}
Context sizes of modern LLMs have grown significantly in recent years. For instance, state-of-the-art models such as GPT-4 and GPT-5 support context lengths ranging from 400K to 1M tokens. However, when applied to lifelong assistant scenarios, such large contexts present several shortcomings.

First, multiple studies~\cite{liu-etal-2024-lost,xu-etal-2022-beyond} have shown that LLMs often perform poorly over long contexts---for example, by neglecting information appearing in the middle of the context content.
Second, processing long contexts is computationally expensive: directly extending transformer architectures to handle longer inputs results in a quadratic increase in both computation time and memory consumption~\cite{memgpt}.
Finally, even with large context windows, user interactions may exceed the fixed context size (e.g., when users share documents). In other words, not all messages in the interaction history 
$H$ can fit into the context.

These limitations have motivated the development of memory-augmented designs~\cite{10.1609/aaai.v38i17.29946,memgpt,modarressi2025memllm,10.1145/3748302,lee-etal-2023-prompted,NEURIPS2025_19909c36}, that store past messages externally. While these systems differ in design details, they share the core idea that any message in the history can be retrieved and incorporated into LLM's context.
For example, Packer et al.~\cite{memgpt} introduced MemGPT, an operating-system-inspired architecture in which an LLM has access to two types of memory: a main context and an external memory. User-LLM interactions are stored in a persistent external datastore, enabling the LLM to retrieve and insert relevant information into its active context when generating a response. Additionally, the LLM maintains a recursive summary of prior conversations along with the most recent exchanges with the user.
We note that these external-memory designs did not consider user-requested deletion of messages from conversation histories.

\section{Problem Definition}
\label{sec:problem}

We consider an LLM-based system with two main goals.
The first goal is utility: to accurately answer the user's queries, given all the information the user has shared with it.
The second goal is to allow the user to delete information from the conversation history~$H$.

The user can request that the LLM deletes information contained in one of the messages in their history $H$.
This could be done either by specifying the identifier $i$ of the message to be
deleted or by providing a predicate query such that all messages matching this query
must be deleted. For example, if a predicate consists of a sequence of keywords, a past message
could be considered a match either based on exact match or semantic similarity. In Figure~\ref{fig:conv},
the first message in the conversation would be the exact match for user's request to remove a sentence ``\emph{I live in Paris}''.

\subsection{Intuition Behind Our Deletion Semantics}
\label{sec:intuition}
As mentioned in the introduction, removing only the message(s) that matches user's request (e.g., deleting semantically close messages) is not sufficient to
remove the information about this message from history $H$. We already saw Example~\ref{ex:1}
in the introduction where the information about user's location is still present in the conversation
after the message explicitly stating the location is removed.
We now consider another example.

\begin{figure}[t]
\begin{center}
\includegraphics[scale=0.4]{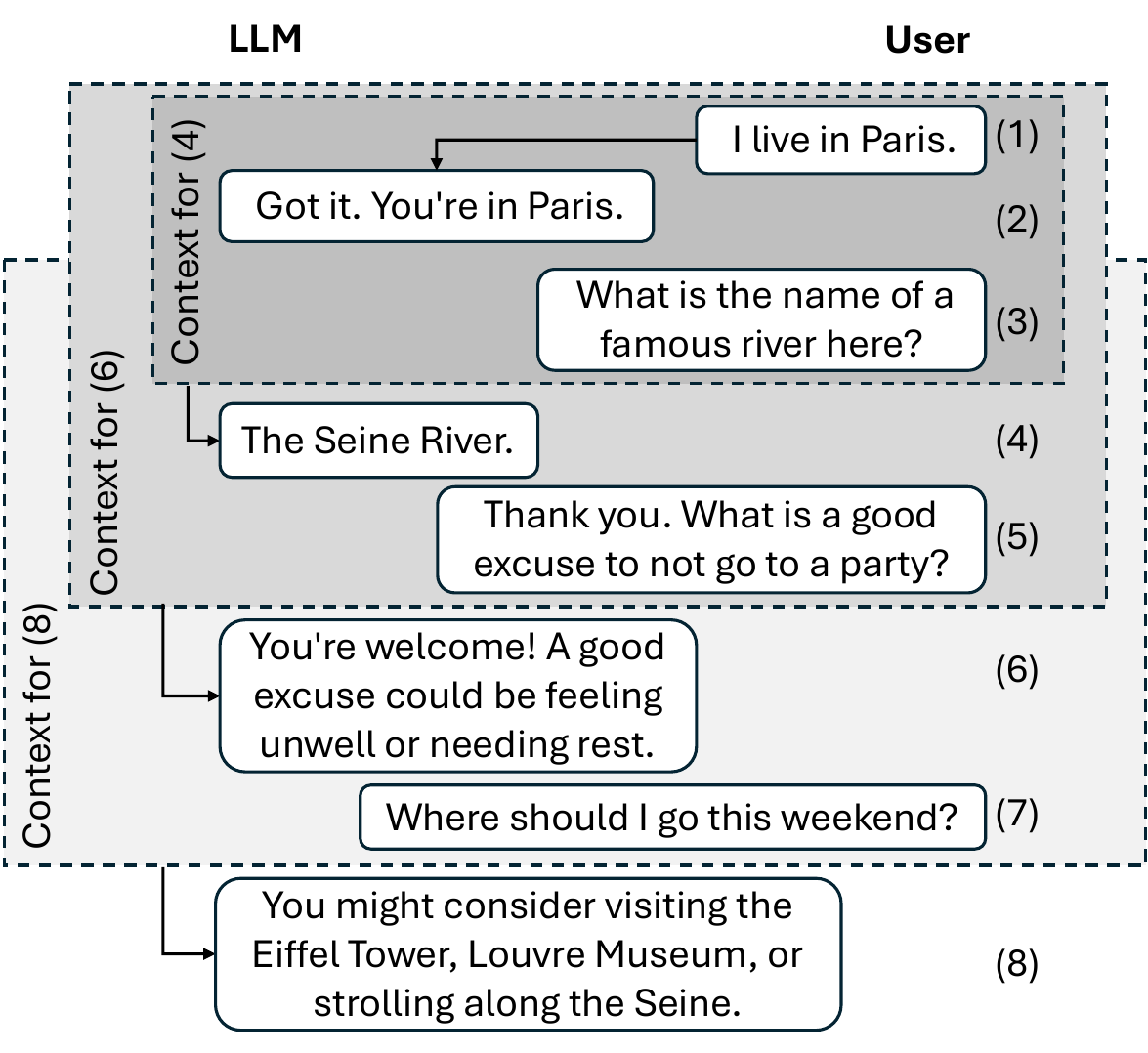}
\caption{
Sliding window context does not guarantee deletion.
Here, an LLM (\texttt{gpt-4.1-nano}) is given as context only the last five messages exchanged between it and the user. Rectangular boundaries
over messages
indicate which messages are included in the contexts of LLM's messages~4, 6 and~8.
 For example, to answer the last user message~7, the LLM is given
messages~3-7. 
Here, even though message 1 is not part of the context of message 8,
information about it persists in message~8 since message~4 depends on 1 and message 4 is part of 8's context.}
\label{fig:interleave}
\end{center}
\end{figure}

\begin{example}
Suppose an LLM is given only the last $l$ messages as part of its context, i.e., context window that
shifts across the history of messages.
Consider an example interaction in Figure~\ref{fig:interleave} where $l=5$. To answer user's
message 7, the LLM is given messages~3-7.
Though this LLM may not be able to reply
to queries that are related to messages exchanged before the last $l$ messages, there is a more serious shortcoming w.r.t.~deletion.

On a surface it may look like the LLM has access only to the last~$l$ interactions and all past messages are automatically deleted from the history.
However, this is not the case: \emph{even if the very first message is no longer part of the context, information about it may still persist in the last $l$ messages}.
This occurs when information contained within a message is propagated across the sliding context window.
In fact, this issue is not specific to the sliding-window setting; it arises whenever the LLM's context contains messages with information derived from earlier messages that are no longer present in the context.

Consider Figure~\ref{fig:interleave} again. Here, message~8 contains information about user's location
even though message 1 was not part of its context. This happened because message~1 was in the context of message 4
and message~4 was in the context of message~8.
\label{ex:3}
\end{example}

Examples~\ref{ex:1} and~\ref{ex:3} demonstrate that deleting an individual message~$m$ is not sufficient; it is also necessary to delete all messages in which $m$ was included in the context.
The main reason is that message~$m$ could have influenced (or tainted) all messages $m'$ where $m \in \context(m')$.
Though this may seem stringent, it guarantees that any messages in which $m$ was part of the context---and were potentially tainted by it---are considered. Whether to delete all of them or only some can be determined based on the application or left to the user's discretion. In this paper, we adopt the strict approach of treating all such messages as potentially containing information about~$m$.

\subsection{Deletion Definition}
\label{sec:deldef}

We now formally define what it means for a message to taint another message and what it means
to delete a message from history~$H$.

\begin{definition}[Message Tainting]
We say that message $m$ has \emph{tainted} (or influenced) LLM's message $m'$ if there is a sequence of messages
$m'_1, m'_2, \ldots, m'_k$ such that:
\begin{eqnarray*}
&m& \in \context(m'_1)\nonumber\\
\wedge\; &m'_i& \in \context(m'_{i+1}),\; \forall_{i=1}^{k-1}{i} \nonumber\\
\wedge\; &m'_k& \in \context(m')
\end{eqnarray*}
\label{def:taint}
\end{definition}
Intuitively, this definition states that there is an information flow from message~$m$ to $m'$.
For example, for $k=0$, all messages in $\context^{-1}(m)$ are tainted by $m$
since $m$ was used directly as part of their context.
Additionally, this definition incorporates all the messages where $m$ appeared indirectly. That is,
all messages that used $\context^{-1}(m)$ in their context are also tainted and so on.

\newcommand{\del}{\mathsf{Dep}}
\newcommand{\Delete}{\mathsf{Delete}}
Since a message can taint more than one message
we define a message dependence set $\del(m)$.

\begin{definition}[Message Dependence Set]
A message dependence set $\del(m)$ is a set of all messages tainted by~message~$m$.
$\del(m)$ is defined recursively as follows:
$$\del(m) = \{m\} \cup \del(m'),\; \forall m' \in \context^{-1}(m)$$
\label{def:depset}
\end{definition}

Note that $\del(m)$ contains $m$, all messages in which $m$ appeared as part of their context, and all messages in which those messages appeared as part of the context, and so on.
The recursion ends upon reaching a message that was not used in the context of any other message.
\begin{example}
In our example in Figure~\ref{fig:conv}, messages 2, 4 and 8 are tainted by message 1, hence,
$\del(m) = \{m_1, m_2, m_4,m_8\}$,
In the example with the shifting window context in Figure~\ref{fig:interleave}, the dependency set is the same using Definition~\ref{def:depset}, since
$m_1 \in \context(m_2)$, $m_2 \in \context(m_4)$ and $m_4 \in \context(m_8)$.
\end{example}

If a set of messages~$M$ needs to be deleted (e.g., in case multiple messages match user's request),
Definition~\ref{def:depset} can be extended to $\del(M)$ and  return $\cup_{m \in M} \del(m)$.

\vspace{5pt}
We are now ready to define what it means to delete $m$ from history $H$.

\begin{definition}[Deletion Function]
Let $\Delete$ be the function that takes as arguments history $H$, a message~$m$ that needs to be removed,
and returns a redacted history~$H'$:
$$H' \leftarrow \Delete(H, m)$$
We say that $\Delete$ removes $m$ from $H$
if $H' \cap  \del(m) = \emptyset$ and $H' \cup  \del(m) = H$, where $\del(m)$ are all messages tainted by $m$ through context inclusion.
\label{def:del}
\end{definition}

The definition states that all messages in  $\del(m)$, which includes~$m$ and all messages tainted by $m$ in $H$, should not be present
in the updated history $H'$. At the same time $H'$ should contain all the other messages from $H$.

We note that deletion from the history does not have to necessarily mean
physical deletion of the corresponding objects in memory. The goal is to ensure that
messages requested to be deleted are not available to the LLM as part of their context.
Besides physical deletion, this could be achieved through marking corresponding messages
as unavailable to be included in certain contexts.

\paragraph{Observations}
Definition~\ref{def:del} may seem stringent, as it mandates deletion of all messages that were tainted by~$m$. We believe that for some applications, this level of strictness may indeed be necessary to ensure that $m$ and all messages influenced by it are deleted. However, there may be applications where this can be relaxed.

For example, one could remove only those messages with~$m$ directly in their context, i.e., consider Definition~\ref{def:taint} with $k=0$, or consider a message as tainted only up to some level $k'>1$. Another approach is to use information-flow-like analysis~\cite{wutschitz2023rethinkingprivacymachinelearning} to determine which messages in the context of~$m'$ have had the most influence on~$m'$. Then, if $m$ was included in $m'$'s context but did not significantly influence it, $m'$ could be excluded from $\del(m)$.

Even for applications where deletion of all dependent messages may not be necessary,
the ability to track and return $\del(m)$ is useful: a user can issue a query to determine
 the potential effect of message deletion and
act accordingly.


\section{Challenges and Design Choices}
\label{sec:design}

We observe that the two goals outlined in the last section---utility and deletion capability---are inherently interleaved.
Satisfying both requires a careful design of which messages are included in the context of an LLM reply.
Consider three examples:

\begin{itemize}
\item Suppose message $m$ is included in the context of many LLM responses.
While this may improve the quality of those responses, it could reduce the utility of future LLM outputs if~$m$ is later deleted,
since $m$ would taint all messages whose context included it (as per Definition~\ref{def:taint}).

\item Including a message in the context of even a single message can still cause many messages to be deleted.
Consider the last message in a conversation: should it be included in the context when generating the next LLM reply?
Suppose it is included, meaning that the LLM's context always contains the current user message $m_i$ along with the last interaction---the user's message $m_{i-2}$ and the LLM's reply $m_{i-1}$.
In this case, deletion of message $m_{i-2}$ requires deletion of messages $m_{i-2}, m_{i-1}, \ldots, m_n$.
That is, following the recursive nature of Definition~\ref{def:taint}, all subsequent messages must be deleted,
since each message has tainted all the following ones.
Including last messages in the context is, however, common in LLMs.

\item Several designs summarize past messages (e.g., via a summary of past conversations and/or user persona facts) and include these summaries as part of the ongoing context~\cite{langchain, memgpt, lee-etal-2023-prompted}.
However, since each summary (even a concise one) is influenced by all the messages used to generate it,
this design suffers from the same issue as including every past message in the context of subsequent messages.
That is, any future message that relies on such a summary becomes tainted by all those prior messages.
As a result, deleting a message from past conversations necessitates deleting all subsequent messages that depended on the tainted summary.
Determining which messages from earlier conversations influenced later ones thus becomes highly non-trivial.

\end{itemize}

These examples raise the following two questions:
\begin{enumerate}
\item Which messages to include in the context of an LLM query to avoid deleting too many (unnecessary) messages?
\item When deleting a message, how to find all LLM messages that have been tainted by it? 
\end{enumerate}

To address these challenges, we propose a design of a system with the following features, which we describe further in~Section~\ref{sec:agent}:

\vspace{5pt}
\emph{\textbf{Managed external storage.}}
Maintaining storage of past user and LLM messages separately allows any past message to be retrieved for use in future queries to an LLM.
This design avoids context-window limitations when relevant messages fall outside of the model's immediate context (see~Section~\ref{sec:longmem}).
Crucially, it also allows us to control which messages are provided to the LLM, ensuring that deleted messages are excluded from future contexts.

\vspace{5pt}
\emph{\textbf{Context construction.}}
For each query, we construct the LLM's context using only the messages most relevant to the current user prompt (e.g., based on semantic similarity).
Restricting the context in this way not only reduces unnecessary downstream deletions, but can also improve performance.
In particular, the LLM can focus on the most pertinent information without being distracted by less relevant content---an advantage also observed in long-memory LLM designs~\cite{liu-etal-2024-lost, xu-etal-2022-beyond}.

\vspace{5pt}
\emph{\textbf{Message provenance.}}
As argued in the introduction and Section~\ref{sec:intuition}, deletion of messages that match (e.g., using cosine similarity) the string  that needs to be removed is insufficient, since messages may reference this information indirectly.
Hence, storing message content alone is insufficient to support faithful deletion.

Let $m$ be a message that must be deleted.
To support deletion as defined in Definition~\ref{def:del}, one must efficiently determine which messages depend on~$m$---that is, identify the dependency set~$\del(m)$ that must also be removed.
In other words, the system must track when a message is used in the context of generating another message.
Explicitly storing this information for all messages and contexts would incur significant memory overhead and require costly searches over message histories.
Instead, we propose an efficient design that represents context dependencies as a directed acyclic graph (DAG), enabling fast identification of message dependencies via graph traversal.

\section{\dellm Framework}
\label{sec:agent}

\dellm aims to satisfy the deletion goal with minimum sacrifice in utility.
To do that it relies on three main components: query processor (\QP), persistent message storage~(\PS) and provenance graph~(\PG).
We briefly describe each component:

\begin{itemize}

\item \QP is an LLM abstraction that takes as input a prompt and responds with a reply. \QP does not have memory of its own and
treats each prompt independently.

 \item \PS is responsible for storing all past conversation messages, user queries and \QP's responses, and supporting efficient retrieval, search and deletion.

\item  \PG is a DAG used to track information flow from
one message to another.
This graph stores
the dependency between conversations and facilitates deletion of all messages related to deletion query including
those tainted by the message to be deleted.

\end{itemize}

\begin{figure}[t]
\begin{center}
\includegraphics[scale=0.5]{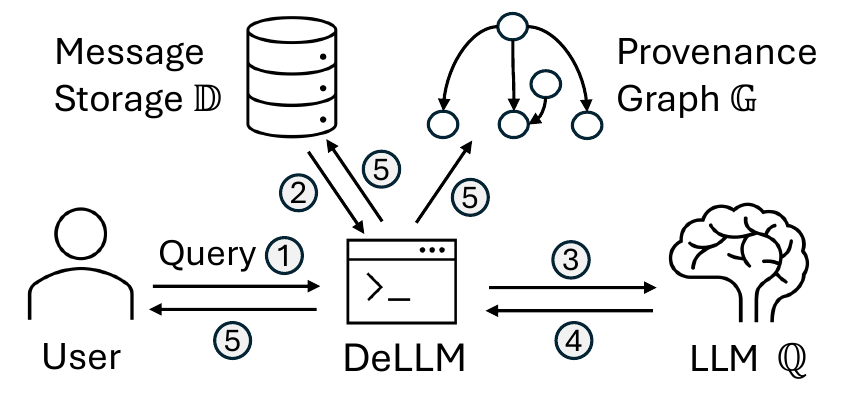}
\caption{Processing a user query with \dellm (Section~\ref{sec:retrieval}):
1)~User sends a query. 2)~\dellm constructs context by retrieving from~\PS messages relevant for answering the query
and~3)~sends the query and the context to LLM. 4)~LLM replies. 5)~\dellm saves
messages in~\PS, adds dependency between LLM's response and messages in its context to~\PG,
and sends the response to the user.}
\label{fig:retrieval}
\end{center}
\end{figure}

\dellm's logic, illustrated in Figures~\ref{fig:retrieval} and~\ref{fig:deletion}, manages the interaction between the three in order to answer user's queries.
Given a user query~$m_{i}$, \dellm's goal is to respond to it by using~\QP and all the messages in the storage.
Since \QP does not have memory of its own, \dellm's task is to generate a context that
will help \QP to answer~$m_i$ while limiting the exposure of all past conversations.
Once the query is answered, \dellm records $m_i$~and~\QP's reply~$m_{i+1}$ in~\PS.
Additionally, it records any dependency between $m_{i+1}$
and past messages that appeared in the context in~\PG.
\PS and \PG are also used to handle deletion requests from the user.

Below we describe in detail the functionality of each component and \dellm's logic when answering user queries.

\subsection{Building Blocks}

\subsubsection{Query Processor}
We abstract an LLM that can handle user queries as a query processor~\QP.
The main requirement, besides utility, is that \QP handles each query
independently and can use only the information available in $m$ and $\context(m)$
to produce an answer.
In order to treat each message independently we require the LLM to not maintain any memory
of past conversations. This can be done by creating a new LLM session for every query.

\QP takes as input a message $m$ and $\context(m)$ that consists of messages relevant to~$m$.
\dellm is responsible for constructing $\context(m)$ of any messages that would help \QP to 
answer $m$.

 We note that even though modern LLMs can take very long contexts (e.g., up to one million tokens),
this complicates tracking of message dependencies. That is,
by default LLM's response would be tainted by all messages given to LLM
in the context (see Section~\ref{sec:design}). Hence, the longer the context, the more messages
will need to be deleted to satisfy Definition~\ref{def:del}.
Hence, we rely on constructing contexts ourselves to be able to control
what appears in the context of every message.
Giving LLM a subset of messages also has utility and performance benefits by allowing LLM
to focus only on relevant messages as opposed to the whole context~\cite{liu-etal-2024-lost,xu-etal-2022-beyond,memgpt}.

\begin{figure}[t]
\begin{center}
\includegraphics[scale=0.5]{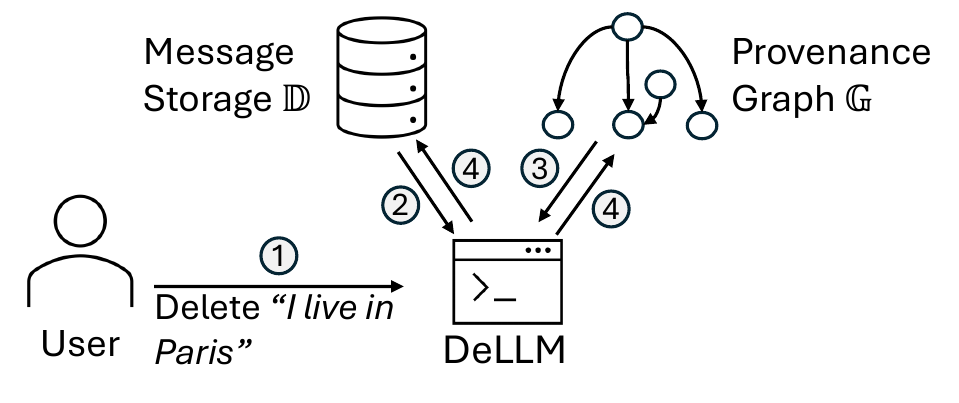}
\caption{Processing a deletion request with \dellm (Section~\ref{sec:deletion}):
1)~User sends a deletion request. 2)~\dellm queries~\PS to retrieve a message set~$M$ matching the deletion criteria.
3)~\dellm traverses~\PG to find messages tainted by messages in~$M$.
4)~\dellm removes $M$ and all the tainted messages from~\PS and~\PG.}
\label{fig:deletion}
\end{center}
\end{figure}

\subsubsection{Message Storage}
Message storage \PS (e.g., database) stores all past messages exchanged between the user and the LLM~\QP.
Storing messages separately is what allows \dellm and other designs to support long memory~\cite{langchaindel,memgpt}.
However, in addition to being able to retrieve past messages, \dellm uses \PS to delete messages as well:
\begin{itemize}
\item
Given a user query $m$, \dellm searches for messages relevant to $m$ in~\PS
and sends them to \QP as part of a query context;
\item If the user requests to delete information, \PS is used to search for messages
that match user's deletion request. These messages are then deleted from \PS. 
\end{itemize}

Each message~$m_i$ is stored alongside its sequential unique identifier~$i$, whether the interaction was user or LLM generated,
and metadata to assist with search. A unique identifier is used to provide a relative order between the messages.

Besides storing messages, \PS supports two queries: search and deletion. Given a phrase, the search functionality returns messages matching a query.
For example, depending on the application, search
can be either based on semantic similarity or keyword matching, or both.
If the search is semantic, a vector embedding is stored as a metadata alongside each message.
For keyword search, a tf-idf-like information (i.e., keyword frequency)  can be associated with each interaction.
An index data structure corresponding to each search technique can be further associated with \PS to speed up the search.

\PS supports deletion of messages via their unique identifiers, via keyword or semantic similarity match to user's query.
In Section~\ref{sec:deletion} we describe \dellm's logic when removing messages.

\subsubsection{Provenance Graph}
\label{sec:pg}

\dellm maintains a directed acyclic graph~\PG to keep track of the information flow between conversation messages.
This ensures that a deletion request removes the requested information and any other
LLM responses that were produced as the result of it (i.e., tainted by it).

Provenance graph \PG contains the following set of nodes and edges.
For every user and LLM message (excluding deletion requests, see Section~\ref{sec:deletion}), \PG 
stores a node. The node name is
the same numeric identifier as the one stored in the persistent storage \PS.
The edges are added to \PG as follows.
Let $m_{i+1}$ be LLM's response to user's message~$m_i$ 
and $\context(m_{i+1})$ be a set of all messages given to the LLM
when replying to $m_i$.
Then \PG records dependency between $m_{i+1}$ and every message in
$\context(m_{i+1})$: for every $m_j \in \context(m_{i+1})$ it adds a directed edge
from $j$ to $i+1$. Since $m_i\in \context(m_{i+1})$, there is also a directed edge from
user's message $i$ to LLM's reply to it, $i+1$.

\begin{figure}[t]
\begin{center}
\includegraphics[scale=0.4]{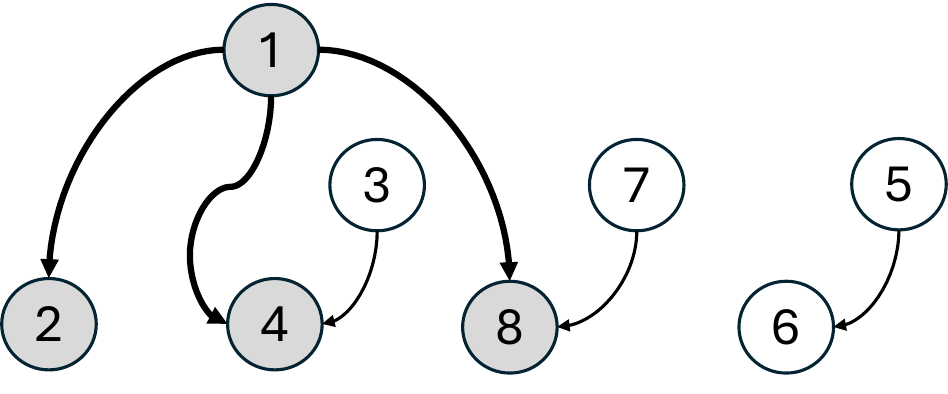}
\caption{A provenance graph,
where a node corresponds to a message identifier and a directed edge from $i$ to $j$
indicates that~$i$ was used as part of a context for message~$j$. If message~1 needs to be removed,
then all messages dependent on~1 need to be removed. They can be found by traversing
the graph to find all the nodes reachable from~1. The traversal will find that messages~2,~4~and~8
need to be removed as well.}
\label{fig:pg}
\end{center}
\end{figure}

  \begin{example}
 Consider the conversation in Figure~\ref{fig:conv}.
 Assume that $\context(m_2) = \{m_1\}$, $\context(m_4) = \{m_1,m_3\}$, $\context(m_6) = \{m_5\}$
 and $\context(m_8) = \{m_1,m_7\}$.
 Then provenance graph for this interaction is illustrated in Figure~\ref{fig:pg}).
 For example, node 8 has two incoming edges to indicate that message $m_8$
 depends on messages~$m_1$ and~$m_7$.
 \end{example}
 
Given a message $m_i$, \dellm needs to find all messages that depend on (or tainted by) it.
That is, we are interested in finding messages in the dependency set $\del(m_i)$ as defined in Definition~\ref{def:depset}.
Graph data structure allows to find this set easily since $\del(m_i)$ is the set of nodes reachable
from~$i$ in \PG. That is, one can find $\del(m_i)$ by traversing \PG starting from~$i$
and recursively following all the outgoing edges from~$i$.

 \begin{example}
 Continuing with the example in Figure~\ref{fig:pg}.
 If message~1 needs to be removed then traversal of \PG from node~1 would return a set \{2,4,8\}.
 \end{example}
 
\PG's size depends on context sizes.
By construction~\PG is a DAG since $\context$ uses only past messages.
Hence, there are no directed edges from~$m_i$ to messages with identifiers less than~$i$.
Let~$s$ be the maximum number of messages in a context. Then, in the worst case,
there are $s$ integers (i.e., message identifiers) stored for every message.
Since $s$ is a constant (e.g., in our experiments it is~5), the graph adds a constant overhead for every stored message
which are often more than a few characters.

\subsection{Answering User Queries}
\label{sec:retrieval}

The user interacts with \dellm by sending messages.
If message $m_i$ is not a deletion request, \dellm
uses \PS to construct context that will help the LLM \QP to answer~$m_i $.
To find messages to include in the context,
\dellm performs a search over data store~\PS given~$m_i$. While we abstract the search functionality to be instantiated based on the use case,
in our implementation we use semantic search (e.g., a message is relevant based
on its cosine similarity score with $m_i$).
All relevant past interactions, that may include last few interactions, serve as a context provided to~\QP to answer~$m_i$.

Given the response $m_{i+1}$ from \QP, \dellm records this interaction as follows.
It writes  $m_i$ and $m_{i+1}$ to \PS and creates a provenance record in~\PG.
In particular, nodes $i$ and $i+1$ are added to~\PG and a directed edge is added from all messages in~$\context(m_{i+1})$ to $m_{i+1}$, including the
user query~$m_i$. 
 \dellm then sends $m_{i+1}$ it to the user.
 (See Figure~\ref{fig:retrieval} for the illustration.)
 
\subsection{Handling Deletion Requests}
\label{sec:deletion}
Some of the messages sent by the user will be deletion requests. The exact format of such requests
can be application specific, for example, \emph{Forget that I live in Paris}, \emph{Delete ``I live in Paris''} or
 \emph{Delete message~1}.
If a user message is indeed a deletion request, \dellm proceeds as follows.
(See Figure~\ref{fig:deletion} for the illustration.)

\dellm extracts the information that needs to be deleted (e.g., \emph{``I live in Paris''} or message with identifier~1) and searches~\PS for relevant messages.
It is up to the application to decide which message(s) match the information the user
asked to remove. For example, it could be an exact match based on message identifier,
an exact or approximate (e.g., semantic) match based on content.
Let $M$ be all the messages that match user's deletion query in~\PS.

\dellm then needs to find all messages dependent on~$M$. Since
\dellm is responsible for constructing LLM's contexts, it knows which messages in $M$ were used in which
contexts. Specifically, this information is stored in~\PG as described in Section~\ref{sec:pg}.
To this end,
\dellm traverses \PG to find all messages reachable from identifiers of messages in~$M$. The union of these messages
will correspond to $\cup \del(m)\; \forall m \in M$ (as per Definition~\ref{def:depset}).

\dellm then removes all the messages in the union from~\PS.
It also removes all the corresponding nodes
and all incoming edges to these nodes from~\QP. Note that there are no outgoing edges from these
messages since such edges would have been traversed and corresponding nodes would be present in the dependency set.
That is, the traversal ends once a node with no outgoing edges is found (i.e., the corresponding message
was not used in a context of any~message).

Different from other user queries a deletion request is not recorded in \PS and \PG. However, it can be if required.
Albeit it should be uniquely marked to ensure it is not returned as part of some message's context, thereby defeating the purpose of deletion.
Finally, as also mentioned in~Section~\ref{sec:deldef}, depending on the application,
messages may not need to be removed from~\PS and~\QP but simply marked as ``removed'' to ensure
they do not used in the future.


\section{Experimental Evaluation}

We implemented  \dellm to evaluate the utility and deletion capability of
our design in practice.
The goal of the evaluation is to investigate the following:

\begin{enumerate}
\item[\textbf{Q1}] The utility impact of  \dellm memory design compared to baseline LLM systems; 
\item[\textbf{Q2}] The deletion capability of  \dellm vs. baselines;
\item[\textbf{Q3}] The impact of deletion on  \dellm's utility;
\item[\textbf{Q4}] The role of provenance graph in deletion.
\end{enumerate}

\subsection{Baselines}
We use OpenAI's \texttt{gpt-4.1-nano} and  \texttt{gpt-5-nano} models as performant cost-efficient LLMs to instantiate the baselines
and  \dellm. These models can process contexts of size 1M and 400K, respectively.
We compare  \dellm to the following baseline LLMs.

 \nomemllm: This LLM configuration does not save any conversations and treats
every interaction with a user from a clean state. That is, the context contains only the user's message
and the instructions.
This represents the best configuration in terms of deletion capability.

\memllm: This system maintains memory in between interactions.
We use LangChain agent design that maintains memory between interactions~\cite{langchain}.

Both baselines are given the following instruction prompt: \emph{``Use the following information from the user, to answer their questions later. Keep your answers very short. If the user asks to remove something remove it from the~conversation.''}

\subsection{ \dellm Implementation Details}
 \dellm is implemented in Python and is $\approx$800 lines of code.
We instantiate LLM \QP using \texttt{gpt-4.1-nano} and  \texttt{gpt-5-nano}.
To instantiate provenance graph~\PG, we use \texttt{rustworkx} Python library to maintain the message dependencies  and perform graph traversal during deletion.
\PS is instantiated using PostgreSQL database that supports keyword and semantic
search. For semantic search we use OpenAI's \texttt{text-embedding-3-small} embeddings of dimension 1,536
and open-source vector similarity search for PostgreSQL, \texttt{pgvector}, based on cosine similarity.

To construct the context for a user prompt,  \dellm retrieves the top 5 messages from~\PS based
on semantic similarity to the user prompt, considering only messages with a cosine similarity score above 0.25 (1 being the perfect match).
Each  query to LLM follows the following context format:
\emph{``Here are past conversations between you and the user}, followed
by the retrieved messages, if any, and \emph{``Reply to the user's query based on past conversation''} followed by
the user prompt.

\subsection{Datasets and Tasks}
\label{sec:data}
We use three datasets to test utility and deletion.
DMR dataset is used to evaluate utility, while PF and TOFU datasets are used to evaluate utility and deletion.
We refer to forget data as data shared by the user that the user will request to remove, and to retain data as the remaining data shared by the user.
PF and TOFU datasets both contain forget and retain data.

\subsubsection{Deep Memory Retrieval (DMR)}
This task was proposed in~\cite{memgpt} based on the Multi-Session Chat (MSC) dataset introduced by Xu et al.~\cite{xu-etal-2022-beyond}.
This is a collection of independent conversations with each one consisting of 5 sessions and each session
containing around 12~messages. Each conversation is generated by two humans acting on behalf of persona
information provided to them. Two unique personas are used for each conversation.
Packer et al.~\cite{memgpt} augmented this dataset with a sixth session containing a question and an answer generated by an LLM.
Each question is asked from the perspective of one of the personas about information in one of the previous five sessions.
The answers are very short and usually consist of one to four words.

The DMR task tests the ability to answer a question about a conversation, given that
gold answers are known. 

We use 100 conversations from this dataset as follows.
\memllm is given instructions and the whole conversation as part of its context since both models we use can
fit the whole conversation.  \nomemllm is given only the question and no context besides the instructions.
In  \dellm, the conversation is stored directly in \PS as it would if it was interactive (albeit w/o information in \PG
as the conversation was not generated by~\dellm).
This dataset is used to test the ability of~ \dellm to answer questions based on a few
messages it retrieves from~\PS as opposed to being given the whole
conversation history. As forget data, we mark the message that corresponds to the answer.

\subsubsection{Persona Facts (PF)}
The DMR dataset consists of fixed conversations in which both user messages and corresponding responses are predefined. To evaluate \dellm's performance in a dynamic setting that more closely reflects real-world user interactions with an LLM, we construct a dataset that contains only user messages, with responses generated dynamically by the LLM during the interaction.

We use an LLM to generate a dataset of 100 records, each consisting of personal facts and questions about them (the exact prompt is provided in~Appendix~\ref{app:prompts}). Specifically, each record contains ten first-person statements describing a persona, followed by five questions or requests that the same person might ask an LLM. These questions should require the LLM to use information from the earlier statements to answer or perform the request properly, without explicitly indicating which statements are relevant (e.g., ``Can you help me plan a weekday routine around my job, commute, and home situation?'').

We construct the forget and retain data as follows. For each persona, an LLM generates four questions about facts contained in the persona statements. One of these facts is designated as information that the user will later request to be removed (i.e., the forget data). The remaining three facts constitute the retain data. For each generated question, the dataset records the gold answer as well as the persona statement containing the information needed to derive that answer.

The task of each LLM system is to answer the four questions based on persona facts shared by the user.
The facts corresponding to the questions are treated as gold responses

We use this dataset as follows. The user sends each piece of information in a persona as a message to an LLM, to which it replies.
To test utility, the LLM is then asked four questions.
To evaluate deletion, the four questions are asked after the systems are requested to remove the corresponding persona statement from the forget dataset.
This helps evaluate not only the deletion rate of the forget data but also utility on the remaining three questions.

\subsubsection{TOFU}
TOFU dataset~\cite{maini2024tofutaskfictitiousunlearning} is a benchmark proposed
to measure unlearning in LLMs. The dataset contains of 200 synthetic author profiles, each given as
20 question-answer pairs. The benchmark contains a split of the profiles into forget and retain sets.
When used for unlearning, the model is first fine-tuned/re-trained
to incorporate the TOFU data, unlearning mechanism is then applied and utility
of responses to questions in forget and retain sets is measured.

We adopt TOFU dataset to our setting as follows.
The answers are given to an LLM as information provided by the user.
The LLM is then prompted on the corresponding questions of forget and retain profiles
to measure the utility before deletion takes place. The original answers in the dataset are treated as gold answers.
To delete information in the forget set, an LLM is requested to remove each
answer in the forget set.
The questions about forget and retain datasets are asked after deletion requests to measure deletion rate and utility after
deletion, correspondingly. The original answers are used as gold responses.

As forget dataset, we choose 1\% of authors in the original dataset, corresponding to 40 facts about two fictitious authors.
As retain dataset, we choose 10\% of other authors, corresponding to 400 facts about 20 fictitious authors, as the retain dataset.
Compared to DMR and PF datasets the information shared by the user (i.e., answers in this case) is longer:
a couple of sentences as opposed to a few words.

\subsection{Metrics}
We evaluate the quality (or utility) of systems' responses against
the gold answers using ROUGE-L recall score~\cite{lin-2004-rouge} and LLM judge.
LLM judge is instructed to evaluate whether
or not the generated response is consistent with the gold
answer, following the same instructions as the ones in~\cite[Section~6.1.2]{memgpt}.
Due to space constraints, we report ROUGE-L scores in Appendix~\ref{app}.
Although they mostly follow trends similar to the accuracy reported by the LLM judge, we observe that the LLM judge evaluates correctness more accurately for answers that do not exactly match the gold answers but convey a similar meaning.

\subsection{Results}

\subsubsection{Utility evaluation (\textbf{Q1})}
 \dellm gives LLM as context a user query and only the most semantically similar messages to it.
To this end, we are interested in evaluating whether \emph{not} providing
all messages in the conversation history impacts the utility of the responses.

Table~\ref{tab:utility} presents accuracy of LLM systems' responses, as measured by LLM judge.
For TOFU, we average utility over forget and retain datasets.
We omit results on the PF dataset since both systems achieve perfect utility on it and, instead, use it to evaluate the impact of deletion.
We observe that on a smaller dataset DMR,
 \dellm loses some accuracy compared to an LLM that is given
all the past messages as part of its context. Instead,  \dellm is given only semantically close
messages, and an answer might be outside of the messages in the context retrieved by  \dellm.
However, for longer messages and longer conversations such as TOFU,
 \dellm performs better than \memllm by 21\% and 2\% on gpt-4.1-nano and on gpt-5-nano, respectively. This is likely due to  \dellm
giving LLM only messages that are semantically close to a user query
as opposed to a long context created by the facts in the forget and retains sets.
Overall, the experiments show that presenting a reduced context is acceptable
and can be even advantageous for long conversations.

\begin{table}[t]
\renewcommand{\arraystretch}{1.2}
\begin{center}
\caption{Utility across datasets, LLM models and systems. Accuracy (\%) based on LLM judge. Higher utility is preferred.}
\begin{tabular}{ r | c | c c}
\multicolumn{1}{c|}{\multirow{2}{*}{LLM}} & \multicolumn{1}{c|}{\multirow{2}{*}{System}}  & \multicolumn{2}{c}{Datasets}\\
\cline{3-4}
& & DMR & TOFU\\
\hline
\multicolumn{1}{c|}{\multirow{2}{*}{gpt-4.1-nano}}
& \memllm & 84\% & 68\%  \\
&  \dellmbf   & 79\% & 89\%  \\
\hline
\multicolumn{1}{c|}{\multirow{2}{*}{gpt-5-nano}}
& \memllm & 76\% & 82\%  \\
&  \dellmbf   & 69\% & 84\%  \\
  \label{tab:utility}
\end{tabular}
\end{center}
\end{table}

\subsubsection{Deletion evaluation (Q2)}
\label{sec:delexp}
We now evaluate the deletion capability of  \dellm and the baselines.
For TOFU and PF datasets, we choose the corresponding forget data (i.e., the gold answers) as the information that the user wishes to remove.
We then compare the utility of answers to questions about deleted facts before and after the deletion requests.
We expect the utility to decrease after the deletion requests are made.
We evaluate deletion on the DMR dataset separately in Section~\ref{sec:abalation} as it contains a pre-defined conversation between two parties.

\begin{table}[t]
\begin{center}
\renewcommand{\arraystretch}{1.2}
\caption{Deletion rate across datasets, LLM models and systems. Accuracy (\%) based on LLM judge.
A lower After score is preferred, as it indicates that the utility on the forget data has decreased following a deletion request for that data.}
\setlength{\tabcolsep}{3pt}
\begin{tabular}{ r | c | r r | r r }
 \multicolumn{1}{c|}{\multirow{3}{*}{LLM}} &
 \multicolumn{1}{c|}{\multirow{3}{*}{System}} &
 \multicolumn{4}{c}{Datasets} \\
\cline{3-6}
 & & \multicolumn{2}{c|}{TOFU} & \multicolumn{2}{c}{PF}\\
\cline{3-6} 
 & & Before & After & Before & After\\
\hline
\multicolumn{1}{c|}{\multirow{4}{*}{gpt-4.1-nano}}
&  \nomemllm & 32.5  & 32.5  & 1  & 1 \\
& \memllm & 75  & 70  & 100  & 96 \\
&  \dellmbf & 95  & 27.5  & 100  & 2.5 \\
\hline
\multicolumn{1}{c|}{\multirow{4}{*}{gpt-5-nano}}
&  \nomemllm & 35  & 35  & 1.7  & 1.7 \\
& \memllm & 89  & 95  & 100  & 98.5 \\
&  \dellmbf & 92.5  & 37.5  & 100  & 2.5 \\

\label{tab:delete}
\end{tabular}
\end{center}
\end{table}

The accuracy on forget data according to LLM judge before and after deletion request is reported in Table~\ref{tab:delete}.
 \nomemllm can be considered as the best case as no other information besides user's
question is known to the LLM.
We observe that \memllm complies with some deletion requests when using the gpt-4.1-nano model,
but largely disregards them with gpt-5-nano. Interestingly, for gpt-5-nano, its utility on the forget data increases after the deletion requests.
One possible explanation is that deletion requests contain the forget data and bring it closer to the subsequent question about it in the conversation context.
In contrast, the before score is measured when questions about the forget data are asked only after the retain facts have been introduced,
making the relevant information more distant in the context.

We observe that  \dellm's deletion rate is much higher than that of \memllm in all cases
and is at most 1.5\% higher than that of \nomemllm.
Upon manual inspection, we observe that there is not information left in \PS containing the answer
and the LLM is making a guess.

\subsubsection{Deletion impact (Q3)}
One way to remove forget data is to remove all the data, even if it is not related to the forget data.
However, this would significantly reduce the utility of the system after the forget request.
To this end, we evaluate whether deleting information related to the forget dataset
affects the utility of the systems on the retain data, i.e., the information that the user shared before the deletion request but
did not request to delete.
We measure the accuracy according to LLM judge on the retain facts before and after forget data
is requested to be deleted.
We expect the utility to remain the same after the deletions were~requested.

The results for \memllm and  \dellm appear in Table~\ref{tab:impact}.
Overall, utility on the retain data is largely unaffected by either system.
We make two observations for the TOFU dataset.
First,  \dellm improves its utility after deletion of forget data.
A possible explanation is that some forget data is no longer retrieved into the relevant context,
reducing interference that previously diluted the useful information.
Second, \memllm achieves lower accuracy on the retain data than on the forget data reported in Table~\ref{tab:delete} (e.g., 62\% vs.~75\%
for gpt-4.1-nano). One possible reason is that the forget data appears earlier in the context than
the retain data and is therefore farther from the user's questions about~it.

\subsubsection{Provenance graph evaluation (Q4)}
\label{sec:abalation}

One way to remove information is by finding all messages that semantically match
the forget data. Though it may work in some circumstances, as we argued in the paper,
this approach can miss LLM messages that refer to forget data indirectly.  \dellm was specifically
designed (1) to find such messages by tracking the dependency of LLM messages on other messages in provenance graph~\PG
and (2) to reduce message dependencies by providing only a few messages as part of LLM's context.
To this end, we now compare deletion rate of semantic deletion approach vs.~\dellm.

\begin{table}[t]
\begin{center}
\renewcommand{\arraystretch}{1.2}
\caption{Impact of deletion of forget data on utility over retain dataset. Accuracy (\%) based on LLM judge.
Higher After score is preferred, as it indicates that the utility on the retain data has not been impacted significantly after the deletion request on forget data.}
\setlength{\tabcolsep}{3pt}
\begin{tabular}{ r | c | r r | r r }
 \multicolumn{1}{c|}{\multirow{3}{*}{LLM}} &
 \multicolumn{1}{c|}{\multirow{3}{*}{System}} &
 \multicolumn{4}{c}{Datasets} \\
\cline{3-6}
& & \multicolumn{2}{c|}{TOFU} & \multicolumn{2}{c}{PF}\\
\cline{3-6}
&& Before & After & Before & After \\
\hline
\multicolumn{1}{c|}{\multirow{2}{*}{gpt-4.1-nano}}
& \memllm & 62  & 62.3  & 100  & 99.7  \\
&  \dellmbf   & 83.5  & 89.3  & 100  & 100  \\
\hline
\multicolumn{1}{c|}{\multirow{2}{*}{gpt-5-nano}}
& \memllm & 74.1  & 74.2  & 100   & 100  \\
&  \dellmbf   & 76.3  & 85.3  & 99.7  & 100 
\label{tab:impact}
\end{tabular}
\end{center}
\end{table}

We use 20 records from PF dataset and evaluate several settings of an LLM with semantic deletion (LLM w/ SD).
This LLM system is given all past messages in order to reply to a user's query.
When it is asked to remove a message, all messages whose similarity score is above a threshold are removed.
Same as for the semantic search used in~ \dellm, we use OpenAI's \texttt{text-embedding-3-small}
to compute cosine similarity.
We use 5  semantic-similarity thresholds  to determine whether a message is a match or not, with higher threshold signifying
higher similarity.

\begin{figure}[t]
\begin{center}
\includegraphics[scale=0.12]{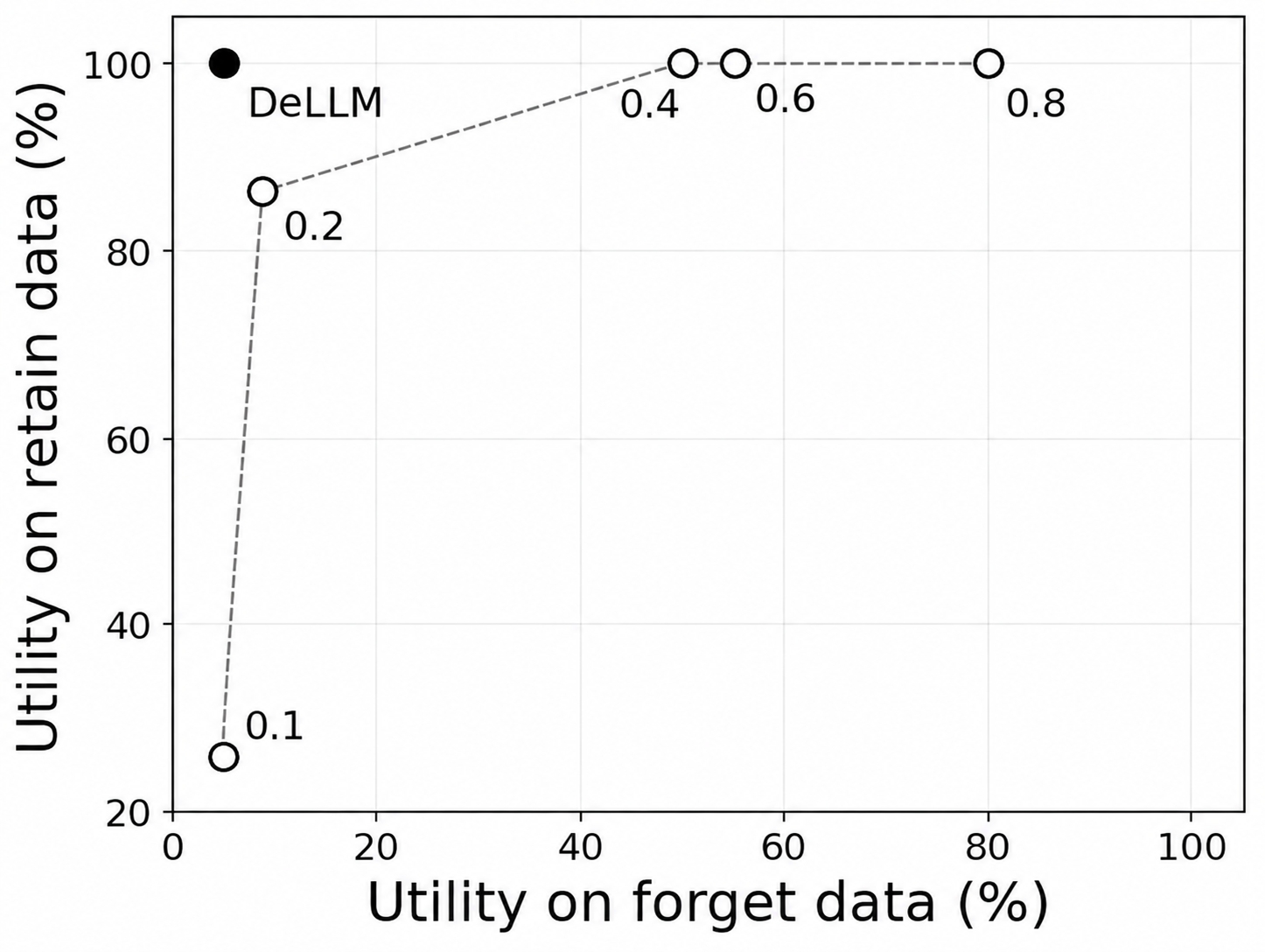}
\caption{Utility on forget and retain data after a deletion request. Lower utility on the forget data and higher utility on the retain data indicate better performance. We compare  \dellm with semantic deletion, evaluated over five similarity thresholds, by removing messages that semantically match the forget data.
To match  \dellm's deletion rate, the semantic deletion method must use a sufficiently aggressive threshold, which results in unrelated messages being deleted and therefore reduces utility on the retain data.
}
\label{fig:semdel}
\end{center}
\end{figure}

As before, we measure systems' utility on the forget and retain data after a deletion request (both systems achieve perfect utility on both datasets before deletion).
We compare  \dellm against several settings of LLM w/ SD~in~Figure~\ref{fig:semdel}.
Both systems use gpt-4.1-nano.

As expected, increasing the semantic-similarity threshold increases the likelihood that forget data is removed, thereby reducing utility on the forget set.
At a threshold of 0.2, semantic deletion reduces forget-set utility to 9\%.
However, this comes at the cost of retain-set utility, which drops from 100\% to 86\%.
To match the deletion rate achieved by  \dellm, LLM w/ SD must remove more messages, including many that are not necessary for satisfying the deletion request, reducing retain-set utility to~25.8\% at 0.1 threshold.

We also evaluate semantic deletion on the DMR dataset, for which the provenance graph is unavailable because the dataset consists of fixed conversations with no explicit information about dependencies between messages. Thus, \dellm cannot be applied beyond deleting messages that match the requested information. Using LLM w/ SD with thresholds of 0.8 and 0.4 results in deletion rates of 40\% and 18\%, respectively, compared with 13\% for \nomemllm.
In contrast, at a thresholds of 0.8 and 0.4, the deletion rates on the above subset of the PF dataset are approximately 80\% and 50\% in Figure~\ref{fig:semdel}. This highlights another limitation of semantic deletion: the appropriate similarity threshold is data-dependent and therefore difficult to choose consistently across datasets.

In summary, compared with \dellm, semantic deletion may either fail to remove messages that retain traces of the ``deleted'' information or remove unrelated messages, thereby degrading overall utility. Effective use of semantic deletion therefore requires careful, dataset-dependent threshold selection and does not provide the same targeted deletion as provenance-aware \dellm.

\section{Discussion}
\label{ref:discussion}

We now discuss several limitations of \dellm and potential improvements that could inspire future work.

\subsection{Ambiguity in deletion requests}
\label{sec:ambiguity}
The deletion requests in our experiments assume that the user specifies which message needs to be removed. Deletion requests could also be expressed in natural language (e.g., ``Remove the information that I live in Paris'').
However, if the request is challenging, recognizing what the user wishes to delete can be challenging.
If the user quotes the phrase that they wish to remove,
then exact match would suffice.
However, a request may rephrase the information that the user wants to remove.
Using semantic or keyword matching may result in deleting information the user did not intend to remove.

To avoid confusion in a deletion request, \dellm can display all the messages it intends to remove and ask for the user's input. These messages will include those that match the deletion request, set $M$, as well as the set of dependent messages, $\del(M)$, along with explanations for why they are candidates for removal (e.g., trace information). It is worth noting that in other settings (e.g., file systems), the user is always prompted to confirm their deletion request. Conversation history could be no different.

Ambiguity could also arise regarding which part of a message needs to be removed. Currently, \dellm adds the user's message to the data store as is, even if the message consists of multiple sentences or paragraphs. As a result, the entire message will be deleted if it matches a deletion request. One could consider splitting the user's messages into, for example, individual sentences and storing each one as a separate message to avoid removing irrelevant parts of a long message. However, as we discuss next, this may complicate dependency analysis between messages.

\subsection{Dependence between user messages}
\label{sec:depend}
\dellm captures dependence between user messages and LLM messages
since it constructs the contexts itself and knows exactly which messages were
used for the context.
However, there maybe dependencies between user messages, either introduced by the user (as discussed in Section~\ref{sec:delexp}) or by splitting long messages
as mentioned above.
For example, ``I live in Paris'' and ``I usually run next to river Seine'' are both related to user's location.
If semantic search is used to identify which messages match user's deletion query (i.e., user's location in this case),
then lowering the threshold on the similarity may result in these messages being added to the pool $M$.
On the flip side, $M$~may contain many false positive and delete too many messages as a result.

\subsection{Conversation summaries}
Several LLM-based systems, including Copilot and ChatGPT, summarize conversation histories or extract facts and user preferences that may be useful in future interactions. As discussed in Section~\ref{sec:design}, directly including such summaries in an LLM's context can lead to excessive deletion: deleting a single message may require deleting an entire summary that used that message (among other messages).
Such summaries can be incorporated into \dellm as follows. Each summary is treated as a derived message, with the provenance graph recording the messages from which it was generated. If any of these source messages is subsequently deleted, the summary can be regenerated from the remaining source messages identified by the provenance graph.
However, summaries should still be selectively included in the LLM's context to limit the number of influenced subsequent messages.

\subsection{Agents and tools} Our setting considered deletion in the presence of LLM, user and a system to manage their communication
to accommodate deletion.
With growing interest in AI Agents and tools we consider how this would impact \dellm.

There are designs~\cite{memgpt,modarressi2025memllm,lee-etal-2023-prompted} where LLM uses tool calls to search external memory instead
of a separate logic, as is the case in~\dellm.
Similarly, one could use an agent with a tool \texttt{delete} that can be called by the LLM itself if it believes that user's message is a deletion request.
However, tool calls may become
subject to prompt injection attacks, asking \dellm to remove information or retrieve too much.
If one does want to use tools, information flow analysis~\cite{zhong2025rtbasdefendingllmagents,costa2025securingaiagentsinformationflow}
could be helpful in determining whether the request is coming from a trusted source. However, as mentioned above, all deletion requests may need to be
confirmed with the user.

Adding agent capabilities to \dellm, for example executing a reasoning loop based on results of tool calls,
is another interesting future direction. For example, one may consider how tool calls, internal and external, and their answers maybe represented in \PS and \PG
and how deletion semantics would need to be extended to tool call responses.


\section{Related work}
\label{sec:related}

\paragraph{Limiting data exposure in agentic systems}
Several works consider protecting the leakage of user data in agentic systems. They aim to prevent private information from being shared in tasks where it is not supposed to appear, either due to adversarial contexts~\cite{10.1145/3658644.3690350} or misunderstandings of the task at hand. For example, one's date of birth may be shared when booking a doctor's appointment but not when booking a restaurant. To mitigate information leakage in inappropriate contexts, methods enforcing privacy policies on the data used in an agent's tasks have been proposed~\cite{10.1145/3658644.3690350,lan2025contextualintegrityllmsreasoning}. A different line of work uses information flow control analysis to determine whether a tool call (e.g., to send an email) originates from a trusted context~\cite{zhong2025rtbasdefendingllmagents,costa2025securingaiagentsinformationflow,wu2024systemleveldefenseindirectprompt} and what type of information---secret or public---it requires for execution. In this paper, we consider a different problem: instead of preventing the leakage of information to untrusted channels, we are interested in deleting information, as well as any other information derived from it, so that it is no longer available in either private or public contexts.

\paragraph{Data deletion in agents}
A recent work by DeChant~\cite{10992571} discusses principles for implementing agentic memories. One of the listed principles is that a user should be able to delete memories. Although this work mentions that one possible way to store memories is via a RAG system~\cite{10.5555/3495724.3496517}, it does not discuss message provenance or its importance in data deletion. It also does not propose a technical solution to implement the suggested principles.

In~\cite{zhong2023memorybank}, the authors compare agent memories to human memories to inspire their memory design. Similar to how humans forget information and inspired by the Ebbinghaus forgetting curve, they introduce a retention rate for past memories based on when they were created and whether they were repeated in subsequent conversations. This contrasts with the approach in this paper, where the user explicitly requests that memories be deleted regardless of when or how many times they were mentioned.

As discussed in the introduction, LLM APIs support deletion at different levels of granularity (e.g., prompts, sessions or facts in conversation summaries). However, this does not allow
one to remove all the dependent and derived information (e.g., as discussed in the introduction, if a message to be deleted was used by the LLM in another session or rephrased). As we argue in this paper,
deleting messages that were derived from the deleted information is essential in ensuring that the content is deleted and does not reappear.

\paragraph*{Machine unlearning}
Machine unlearning focuses on removing the influence of a training data point on model's parameters and predictions~\cite{10.1109/SP.2015.35}. There has been extensive work on unlearning techniques, ranging from exact but expensive methods, in which a model---or parts of it---is retrained without the forget data~\cite{9519428}, to approximate techniques, where model parameters are updated~\cite{patil2024can,chen2023unlearn,eldan2023whosharrypotterapproximate}
based on the estimated influence of the forget data. Although approximate unlearning is more efficient than exact unlearning, it has been shown that one may still be able to extract forget data from the updated model~\cite{maini2024tofutaskfictitiousunlearning,10992375}.

Our work addresses a problem distinct from machine unlearning and departs from it both conceptually and technically. First, we consider the deletion of memories created through interactions between an LLM and a user, rather than the deletion of data used to train the LLM. To this end, we propose a definition of deletion specific to this setting: it considers both the deletion of content matching a deletion predicate and the removal of all consequent messages derived from this content. Moreover, our technique is based on tracking context dependencies to ensure that data derived from the deleted content is also removed.

\paragraph{Data provenance}
Provenance is defined as the place of origin of an object. For a digital object, provenance includes a description of how the object was derived~\cite{10.1007/3-540-44503-X_20}. For example, in software development, build rules express provenance, and source code control systems track it. There has been a long line of work on automatically tracking provenance in file systems~\cite{10.5555/1267359.1267363} and databases~\cite{ilprints658,10.1007/3-540-44503-X_20}. Deletion of data was considered in~\cite{ilprints658}, where deleted data is not removed but instead becomes expired. As mentioned in Section~\ref{sec:problem}, this is an approach that \dellm can adopt as well.
To our knowledge, our work is the first to suggest maintaining provenance of messages in LLM conversations and to use it to support faithful deletion.
 
\section{Conclusion}
In this paper, we posed the problem of deleting messages from persistent LLM memories. Our definition considers deletion to be correct if a message, along with all messages derived from it by the LLM, is removed. Since current LLMs and LLM memory designs do not support this level of deletion, we proposed \dellm. \dellm acts as a proxy between a user and an LLM, responsible for storing and retrieving messages from external storage to help the LLM answer queries, and deleting this messages
as per user request. A crucial part of our design is the provenance graph, which maintains message dependencies and allows \dellm to identify messages derived from a message that needs to be deleted. This ensures that traces of the information are removed from the message history. Our evaluation demonstrates the tradeoff between deletion and utility achieved by \dellm compared to baselines that have access to all memories or no memories at all.

\section*{Acknowledgment}

The author worked on this project in part while visiting Azure Research -- Security and Privacy group at Microsoft Cambridge during her sabbatical,
which was partially funded by Microsoft and the Faculty of Engineering and Information Technology at the University of Melbourne. She is deeply grateful to Boris K\"{o}pf and Santiago Zanella-B\'{e}guelin for the insightful discussions about this work and for their helpful comments on the initial drafts.
The author would also like to thank Manuel Costa, C\'{e}dric Fournet, Shruti Tople and Lukas Wutschitz for valuable discussions
during her visit.

\bibliographystyle{IEEEtranS}

\begin{thebibliography}{10}
\providecommand{\url}[1]{#1}
\csname url@samestyle\endcsname
\providecommand{\newblock}{\relax}
\providecommand{\bibinfo}[2]{#2}
\providecommand{\BIBentrySTDinterwordspacing}{\spaceskip=0pt\relax}
\providecommand{\BIBentryALTinterwordstretchfactor}{4}
\providecommand{\BIBentryALTinterwordspacing}{\spaceskip=\fontdimen2\font plus
\BIBentryALTinterwordstretchfactor\fontdimen3\font minus
  \fontdimen4\font\relax}
\providecommand{\BIBforeignlanguage}[2]{{%
\expandafter\ifx\csname l@#1\endcsname\relax
\typeout{** WARNING: IEEEtranS.bst: No hyphenation pattern has been}%
\typeout{** loaded for the language `#1'. Using the pattern for}%
\typeout{** the default language instead.}%
\else
\language=\csname l@#1\endcsname
\fi
#2}}
\providecommand{\BIBdecl}{\relax}
\BIBdecl

\bibitem{amazondel}
\BIBentryALTinterwordspacing
Amazon, ``{DeleteAgentMemory},'' 2026, accessed in August 2026. [Online].
  Available:
  \url{https://docs.aws.amazon.com/bedrock/latest/APIReference/API\_agent-runtime\_DeleteAgentMemory.html}
\BIBentrySTDinterwordspacing

\bibitem{claude-memory}
\BIBentryALTinterwordspacing
Anthropic, ``Bringing memory to teams at work,'' 2026, accessed in August 2026.
  [Online]. Available: \url{https://www.anthropic.com/news/memory}
\BIBentrySTDinterwordspacing

\bibitem{10.1145/3658644.3690350}
\BIBentryALTinterwordspacing
E.~Bagdasarian, R.~Yi, S.~Ghalebikesabi, P.~Kairouz, M.~Gruteser, S.~Oh,
  B.~Balle, and D.~Ramage, ``{AirGapAgent}: Protecting privacy-conscious
  conversational agents,'' in \emph{Proceedings of the 2024 on ACM SIGSAC
  Conference on Computer and Communications Security}, ser. CCS '24.\hskip 1em
  plus 0.5em minus 0.4em\relax New York, NY, USA: Association for Computing
  Machinery, 2024, pp. 3868--3882. [Online]. Available:
  \url{https://doi.org/10.1145/3658644.3690350}
\BIBentrySTDinterwordspacing

\bibitem{9519428}
L.~Bourtoule, V.~Chandrasekaran, C.~A. Choquette-Choo, H.~Jia, A.~Travers,
  B.~Zhang, D.~Lie, and N.~Papernot, ``Machine unlearning,'' in \emph{2021 IEEE
  Symposium on Security and Privacy (SP)}, 2021, pp. 141--159.

\bibitem{10.1007/3-540-44503-X_20}
P.~Buneman, S.~Khanna, and T.~Wang-Chiew, ``Why and where: A characterization
  of data provenance,'' in \emph{Database Theory --- ICDT 2001}, J.~Van~den
  Bussche and V.~Vianu, Eds.\hskip 1em plus 0.5em minus 0.4em\relax Berlin,
  Heidelberg: Springer Berlin Heidelberg, 2001, pp. 316--330.

\bibitem{10.1109/SP.2015.35}
\BIBentryALTinterwordspacing
Y.~Cao and J.~Yang, ``Towards making systems forget with machine unlearning,''
  in \emph{Proceedings of the 2015 IEEE Symposium on Security and Privacy},
  ser. SP '15.\hskip 1em plus 0.5em minus 0.4em\relax USA: IEEE Computer
  Society, 2015, pp. 463--480. [Online]. Available:
  \url{https://doi.org/10.1109/SP.2015.35}
\BIBentrySTDinterwordspacing

\bibitem{chen2023unlearn}
\BIBentryALTinterwordspacing
J.~Chen and D.~Yang, ``Unlearn what you want to forget: Efficient unlearning
  for {LLM}s,'' in \emph{The 2023 Conference on Empirical Methods in Natural
  Language Processing}, 2023. [Online]. Available:
  \url{https://openreview.net/forum?id=GprvtTwOxy}
\BIBentrySTDinterwordspacing

\bibitem{costa2025securingaiagentsinformationflow}
\BIBentryALTinterwordspacing
M.~Costa, B.~K\"{o}pf, A.~Kolluri, A.~Paverd, M.~Russinovich, A.~Salem,
  S.~Tople, L.~Wutschitz, and S.~Zanella-B\`{e}guelin, ``Securing ai agents
  with information-flow control,'' 2025. [Online]. Available:
  \url{https://arxiv.org/abs/2505.23643}
\BIBentrySTDinterwordspacing

\bibitem{10992571}
\BIBentryALTinterwordspacing
C.~DeChant, ``{ Episodic Memory in AI Agents Poses Risks that Should be Studied
  and Mitigated },'' in \emph{2025 IEEE Conference on Secure and Trustworthy
  Machine Learning (SaTML)}.\hskip 1em plus 0.5em minus 0.4em\relax Los
  Alamitos, CA, USA: IEEE Computer Society, Apr. 2025, pp. 321--332. [Online].
  Available:
  \url{https://doi.ieeecomputersociety.org/10.1109/SaTML64287.2025.00024}
\BIBentrySTDinterwordspacing

\bibitem{eldan2023whosharrypotterapproximate}
\BIBentryALTinterwordspacing
R.~Eldan and M.~Russinovich, ``Who's harry potter? approximate unlearning in
  llms,'' 2023. [Online]. Available: \url{https://arxiv.org/abs/2310.02238}
\BIBentrySTDinterwordspacing

\bibitem{gemini-memory}
\BIBentryALTinterwordspacing
Google, ``{Get personalization with memory of your past Gemini chats},'' 2026,
  accessed in August 2026. [Online]. Available:
  \url{https://support.google.com/gemini/answer/16598623}
\BIBentrySTDinterwordspacing

\bibitem{gemini}
\BIBentryALTinterwordspacing
------, ``{Introducing Gemini, your new personal AI assistant},'' 2026,
  accessed in August 2026. [Online]. Available:
  \url{https://gemini.google/us/assistant}
\BIBentrySTDinterwordspacing

\bibitem{10.1145/3605764.3623985}
\BIBentryALTinterwordspacing
K.~Greshake, S.~Abdelnabi, S.~Mishra, C.~Endres, T.~Holz, and M.~Fritz, ``Not
  what you've signed up for: Compromising real-world {LLM}-integrated
  applications with indirect prompt injection,'' in \emph{Proceedings of the
  16th ACM Workshop on Artificial Intelligence and Security}, ser. AISec
  '23.\hskip 1em plus 0.5em minus 0.4em\relax New York, NY, USA: Association
  for Computing Machinery, 2023, pp. 79--90. [Online]. Available:
  \url{https://doi.org/10.1145/3605764.3623985}
\BIBentrySTDinterwordspacing

\bibitem{10992375}
\BIBentryALTinterwordspacing
J.~Hayes, I.~Shumailov, E.~Triantafillou, A.~Khalifa, and N.~Papernot,
  ``{Inexact Unlearning Needs More Careful Evaluations to Avoid a False Sense
  of Privacy},'' in \emph{2025 IEEE Conference on Secure and Trustworthy
  Machine Learning (SaTML)}.\hskip 1em plus 0.5em minus 0.4em\relax Los
  Alamitos, CA, USA: IEEE Computer Society, Apr. 2025, pp. 497--519. [Online].
  Available:
  \url{https://doi.ieeecomputersociety.org/10.1109/SaTML64287.2025.00034}
\BIBentrySTDinterwordspacing

\bibitem{10.1145/3706598.3713819}
\BIBentryALTinterwordspacing
M.~J\"{o}rke, S.~Sapkota, L.~Warkenthien, N.~Vainio, P.~Schmiedmayer,
  E.~Brunskill, and J.~A. Landay, ``{GPTCoach: Towards LLM-Based Physical
  Activity Coaching},'' in \emph{Proceedings of the 2025 CHI Conference on
  Human Factors in Computing Systems}, ser. CHI '25.\hskip 1em plus 0.5em minus
  0.4em\relax New York, NY, USA: Association for Computing Machinery, 2025.
  [Online]. Available: \url{https://doi.org/10.1145/3706598.3713819}
\BIBentrySTDinterwordspacing

\bibitem{lan2025contextualintegrityllmsreasoning}
G.~Lan, H.~A. Inan, S.~Abdelnabi, J.~Kulkarni, L.~Wutschitz, R.~Shokri, C.~G.
  Brinton, and R.~Sim, ``Contextual integrity in {LLMs} via reasoning and
  reinforcement learning,'' in \emph{To appear in Proceedings of the 39th
  International Conference on Neural Information Processing Systems}, 2025.

\bibitem{langchain}
\BIBentryALTinterwordspacing
LangChain, ``Agents,'' 2026, accessed in August 2026. [Online]. Available:
  \url{https://docs.langchain.com/oss/python/langchain/agents}
\BIBentrySTDinterwordspacing

\bibitem{langchaindel}
\BIBentryALTinterwordspacing
------, ``{Lang-MemGPT},'' 2026, accessed in August 2026. [Online]. Available:
  \url{https://github.com/langchain-ai/lang-memgpt}
\BIBentrySTDinterwordspacing

\bibitem{lee-etal-2023-prompted}
\BIBentryALTinterwordspacing
G.~Lee, V.~Hartmann, J.~Park, D.~Papailiopoulos, and K.~Lee, ``Prompted {LLM}s
  as chatbot modules for long open-domain conversation,'' in \emph{Findings of
  the Association for Computational Linguistics: ACL 2023}, A.~Rogers,
  J.~Boyd-Graber, and N.~Okazaki, Eds.\hskip 1em plus 0.5em minus 0.4em\relax
  Toronto, Canada: Association for Computational Linguistics, Jul. 2023, pp.
  4536--4554. [Online]. Available:
  \url{https://aclanthology.org/2023.findings-acl.277/}
\BIBentrySTDinterwordspacing

\bibitem{10.5555/3495724.3496517}
P.~Lewis, E.~Perez, A.~Piktus, F.~Petroni, V.~Karpukhin, N.~Goyal,
  H.~K\"{u}ttler, M.~Lewis, W.-t. Yih, T.~Rockt\"{a}schel, S.~Riedel, and
  D.~Kiela, ``Retrieval-augmented generation for knowledge-intensive nlp
  tasks,'' in \emph{Proceedings of the 34th International Conference on Neural
  Information Processing Systems}, ser. NIPS '20.\hskip 1em plus 0.5em minus
  0.4em\relax Red Hook, NY, USA: Curran Associates Inc., 2020.

\bibitem{lin-2004-rouge}
\BIBentryALTinterwordspacing
C.-Y. Lin, ``{ROUGE}: A package for automatic evaluation of summaries,'' in
  \emph{Text Summarization Branches Out}.\hskip 1em plus 0.5em minus
  0.4em\relax Barcelona, Spain: Association for Computational Linguistics, Jul.
  2004, pp. 74--81. [Online]. Available:
  \url{https://aclanthology.org/W04-1013/}
\BIBentrySTDinterwordspacing

\bibitem{liu-etal-2024-lost}
\BIBentryALTinterwordspacing
N.~F. Liu, K.~Lin, J.~Hewitt, A.~Paranjape, M.~Bevilacqua, F.~Petroni, and
  P.~Liang, ``Lost in the middle: How language models use long contexts,''
  \emph{Transactions of the Association for Computational Linguistics},
  vol.~12, pp. 157--173, 2024. [Online]. Available:
  \url{https://aclanthology.org/2024.tacl-1.9/}
\BIBentrySTDinterwordspacing

\bibitem{maini2024tofutaskfictitiousunlearning}
\BIBentryALTinterwordspacing
P.~Maini, Z.~Feng, A.~Schwarzschild, Z.~C. Lipton, and J.~Z. Kolter, ``{TOFU: A
  Task of Fictitious Unlearning for LLMs},'' 2024. [Online]. Available:
  \url{https://arxiv.org/abs/2401.06121}
\BIBentrySTDinterwordspacing

\bibitem{copilot}
\BIBentryALTinterwordspacing
Microsoft, ``{AI agents are changing the way we work},'' 2026, accessed in
  August 2026. [Online]. Available:
  \url{https://www.microsoft.com/en-us/microsoft-365-copilot/agents}
\BIBentrySTDinterwordspacing

\bibitem{copilot-memory}
\BIBentryALTinterwordspacing
------, ``{Conversation history in Microsoft Copilot},'' 2026, accessed in
  August 2026. [Online]. Available:
  \url{https://support.microsoft.com/en-us/microsoft-copilot/conversation-history-in-microsoft-copilot}
\BIBentrySTDinterwordspacing

\bibitem{copilot-managemem}
\BIBentryALTinterwordspacing
------, ``{Manage Copilot Memory in Microsoft Copilot},'' 2026, accessed in
  August 2026. [Online]. Available:
  \url{https://support.microsoft.com/en-us/microsoft-365-copilot/manage-copilot-memory-in-microsoft-365-copilot}
\BIBentrySTDinterwordspacing

\bibitem{copilot-summary}
\BIBentryALTinterwordspacing
------, ``{Personalize what Microsoft Copilot remembers},'' 2026, accessed in
  August 2026. [Online]. Available:
  \url{https://support.microsoft.com/en-us/microsoft-365-copilot/personalize-what-microsoft-365-copilot-remembers}
\BIBentrySTDinterwordspacing

\bibitem{modarressi2025memllm}
\BIBentryALTinterwordspacing
A.~Modarressi, A.~K{\"o}ksal, A.~Imani, M.~Fayyaz, and H.~Schuetze, ``Mem{LLM}:
  Finetuning {LLM}s to use explicit read-write memory,'' \emph{Transactions on
  Machine Learning Research}, 2025. [Online]. Available:
  \url{https://openreview.net/forum?id=dghM7sOudh}
\BIBentrySTDinterwordspacing

\bibitem{10.5555/1267359.1267363}
K.-K. Muniswamy-Reddy, D.~A. Holland, U.~Braun, and M.~Seltzer,
  ``Provenance-aware storage systems,'' in \emph{Proceedings of the Annual
  Conference on USENIX '06 Annual Technical Conference}, ser. ATEC '06.\hskip
  1em plus 0.5em minus 0.4em\relax USA: USENIX Association, 2006, p.~4.

\bibitem{openai-booking}
\BIBentryALTinterwordspacing
OpenAI, ``{Booking.com and OpenAI personalize travel at scale},'' 2026,
  accessed in August 2026. [Online]. Available:
  \url{https://openai.com/index/booking-com/}
\BIBentrySTDinterwordspacing

\bibitem{chatgpt-memory}
\BIBentryALTinterwordspacing
------, ``{Dreaming: Better memory for a more helpful ChatGPT},'' 2026,
  accessed in August 2026. [Online]. Available:
  \url{https://openai.com/index/chatgpt-memory-dreaming/}
\BIBentrySTDinterwordspacing

\bibitem{chatgpt-memoryfaq}
\BIBentryALTinterwordspacing
------, ``{How memory works},'' 2026, accessed in August 2026. [Online].
  Available:
  \url{https://help.openai.com/en/articles/8590148-memory-in-chatgpt}
\BIBentrySTDinterwordspacing

\bibitem{openai-sdk}
\BIBentryALTinterwordspacing
------, ``Openai agents sdk,'' 2026, accessed in August 2026. [Online].
  Available: \url{https://github.com/openai/openai-agents-python}
\BIBentrySTDinterwordspacing

\bibitem{memgpt}
C.~Packer, S.~Wooders, K.~Lin, V.~Fang, S.~G. Patil, I.~Stoica, and J.~E.
  Gonzalez, ``{MemGPT}: Towards {LLMs} as operating systems,'' \emph{arXiv
  preprint arXiv:2310.08560}, 2023.

\bibitem{patil2024can}
\BIBentryALTinterwordspacing
V.~Patil, P.~Hase, and M.~Bansal, ``Can sensitive information be deleted from
  {LLM}s? objectives for defending against extraction attacks,'' in \emph{The
  Twelfth International Conference on Learning Representations}, 2024.
  [Online]. Available: \url{https://openreview.net/forum?id=7erlRDoaV8}
\BIBentrySTDinterwordspacing

\bibitem{ilprints658}
\BIBentryALTinterwordspacing
J.~Widom, ``Trio: A system for integrated management of data, accuracy, and
  lineage,'' Stanford InfoLab, Technical Report 2004-40, August 2004. [Online].
  Available: \url{http://ilpubs.stanford.edu:8090/658/}
\BIBentrySTDinterwordspacing

\bibitem{wu2024systemleveldefenseindirectprompt}
\BIBentryALTinterwordspacing
F.~Wu, E.~Cecchetti, and C.~Xiao, ``System-level defense against indirect
  prompt injection attacks: An information flow control perspective,'' 2024.
  [Online]. Available: \url{https://arxiv.org/abs/2409.19091}
\BIBentrySTDinterwordspacing

\bibitem{wutschitz2023rethinkingprivacymachinelearning}
\BIBentryALTinterwordspacing
L.~Wutschitz, B.~K\"{o}pf, A.~Paverd, S.~Rajmohan, A.~Salem, S.~Tople,
  S.~Zanella-B\`{e}guelin, M.~Xia, and V.~R\"{u}hle, ``Rethinking privacy in
  machine learning pipelines from an information flow control perspective,''
  2023. [Online]. Available: \url{https://arxiv.org/abs/2311.15792}
\BIBentrySTDinterwordspacing

\bibitem{xu-etal-2022-beyond}
\BIBentryALTinterwordspacing
J.~Xu, A.~Szlam, and J.~Weston, ``Beyond goldfish memory: Long-term open-domain
  conversation,'' in \emph{Proceedings of the 60th Annual Meeting of the
  Association for Computational Linguistics (Volume 1: Long Papers)},
  S.~Muresan, P.~Nakov, and A.~Villavicencio, Eds.\hskip 1em plus 0.5em minus
  0.4em\relax Dublin, Ireland: Association for Computational Linguistics, May
  2022, pp. 5180--5197. [Online]. Available:
  \url{https://aclanthology.org/2022.acl-long.356/}
\BIBentrySTDinterwordspacing

\bibitem{NEURIPS2025_19909c36}
\BIBentryALTinterwordspacing
W.~Xu, Z.~Liang, K.~Mei, H.~Gao, J.~Tan, and Y.~Zhang, ``A-mem: Agentic memory
  for llm agents,'' in \emph{Advances in Neural Information Processing
  Systems}, D.~Belgrave, C.~Zhang, H.~Lin, R.~Pascanu, P.~Koniusz, M.~Ghassemi,
  and N.~Chen, Eds., vol. 38, Main Conference.\hskip 1em plus 0.5em minus
  0.4em\relax Curran Associates, Inc., 2025, pp. 17\,577--17\,604. [Online].
  Available:
  \url{https://proceedings.neurips.cc/paper_files/paper/2025/file/19909c36f51abc4856b4560aff3d36d6-Paper-Conference.pdf}
\BIBentrySTDinterwordspacing

\bibitem{10.1145/3748302}
\BIBentryALTinterwordspacing
Z.~Zhang, Q.~Dai, X.~Bo, C.~Ma, R.~Li, X.~Chen, J.~Zhu, Z.~Dong, and J.-R. Wen,
  ``A survey on the memory mechanism of large language model-based agents,''
  \emph{ACM Trans. Inf. Syst.}, vol.~43, no.~6, Sep. 2025. [Online]. Available:
  \url{https://doi.org/10.1145/3748302}
\BIBentrySTDinterwordspacing

\bibitem{zhong2025rtbasdefendingllmagents}
\BIBentryALTinterwordspacing
P.~Y. Zhong, S.~Chen, R.~Wang, M.~McCall, B.~L. Titzer, H.~Miller, and P.~B.
  Gibbons, ``{RTBAS: Defending LLM Agents Against Prompt Injection and Privacy
  Leakage},'' 2025. [Online]. Available: \url{https://arxiv.org/abs/2502.08966}
\BIBentrySTDinterwordspacing

\bibitem{zhong2023memorybank}
W.~Zhong, L.~Guo, Q.~Gao, and Y.~Wang, ``Memorybank: Enhancing large language
  models with long-term memory,'' \emph{arXiv preprint arXiv:2305.10250}, 2023.

\bibitem{10.1609/aaai.v38i17.29946}
\BIBentryALTinterwordspacing
W.~Zhong, L.~Guo, Q.~Gao, H.~Ye, and Y.~Wang, ``{MemoryBank}: enhancing large
  language models with long-term memory,'' in \emph{Proceedings of the
  Thirty-Eighth AAAI Conference on Artificial Intelligence and Thirty-Sixth
  Conference on Innovative Applications of Artificial Intelligence and
  Fourteenth Symposium on Educational Advances in Artificial Intelligence},
  ser. AAAI'24/IAAI'24/EAAI'24.\hskip 1em plus 0.5em minus 0.4em\relax AAAI
  Press, 2024. [Online]. Available:
  \url{https://doi.org/10.1609/aaai.v38i17.29946}
\BIBentrySTDinterwordspacing

\end{thebibliography}


\appendix

\subsection{Examples}

All examples in the paper were completed in August 2026 using the following models: Gemini 3.6 Flash; Microsoft 365 Copilot, based on the GPT-5 chat model; and ChatGPT, powered by GPT-5.6 Sol.

\begin{figure*}[t]
    \centering
    \begin{subfigure}[t]{0.48\textwidth}
    \vspace{0pt}
        \centering
        \includegraphics[width=\linewidth]{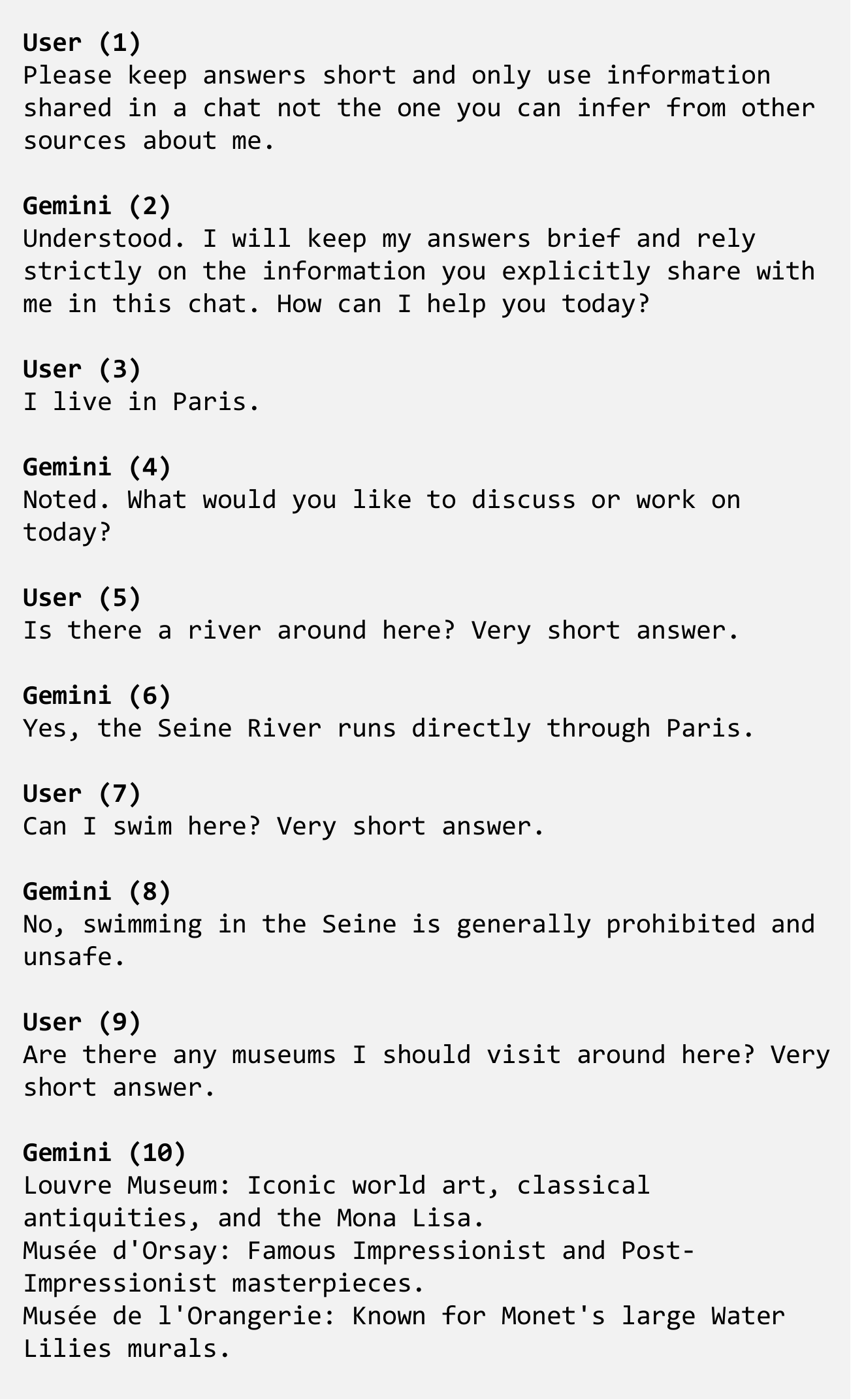}
    \end{subfigure}
    \hfill
    \begin{subfigure}[t]{0.48\textwidth}
    \vspace{0pt}
        \centering
        \includegraphics[width=\linewidth]{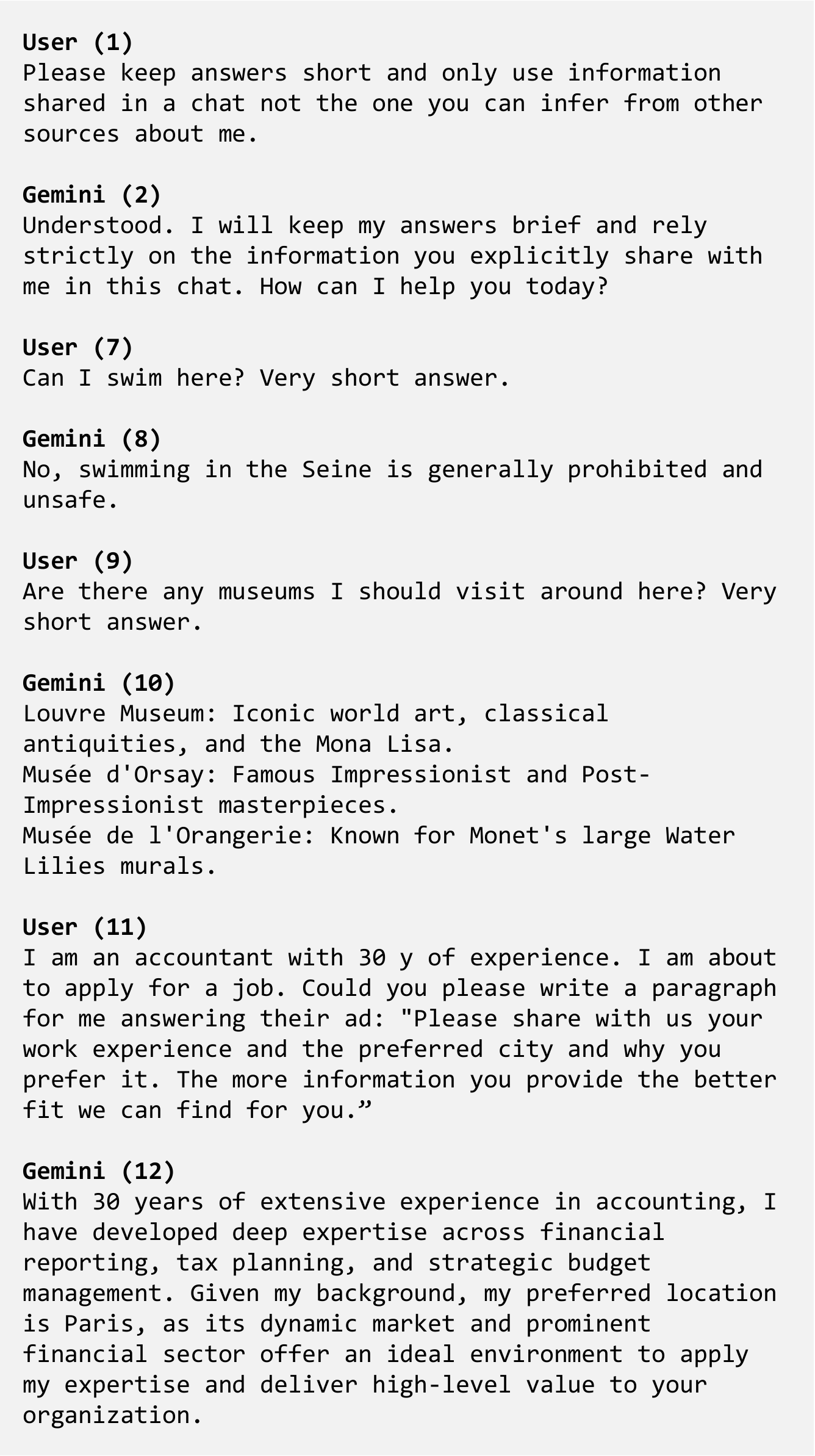}
    \end{subfigure}
    \caption{(Left) Sample interaction with Gemini where the user shares
their location and asks several questions that depend on the location. (Right) The user deletes messages 3--6 that mention the location ``Paris'' explicitly in the conversation in the left figure. The user then continues the conversation by asking question (11). Gemini's response still uses the location ``Paris'' that is inferred from
the remaining messages. Note: This interaction happened outside of France and the authors are not based in France.}
\label{fig:gemini}
\end{figure*}

Figure~\ref{fig:gemini} shows an example where deleting individual prompts that explicitly mention a piece of information is insufficient to remove it, because the remaining messages can still refer to that information indirectly.

\subsection{ROUGE-L scores}

We include ROUGE-L recall scores in Tables~\ref{tab:utility-r},~\ref{tab:delete-r} and~\ref{tab:impact-r} for the same experiments
as those reported in Tables~\ref{tab:utility},~\ref{tab:delete} and~\ref{tab:impact}, correspondingly, with accuracy based on LLM judge.

We observe that LLM judge measures correctness more accurately for answers
that do not match gold answers exactly but have a similar meaning.
For example, given the following question:
\emph{``What is the full name of the author born in Kuwait City, Kuwait on 08/09/1956?''}
 and its gold answer \emph{``The full name of the fictitious author born in Kuwait City, Kuwait on the 8th of September, 1956 is Basil Mahfouz Al-Kuwaiti.''},
the LLM judge gives \emph{False} to \nomemllm's answer \emph{``I'm missing the author's name from the information provided. Could you share the author's works or any other detail to identify them?''} and \emph{True} to \dellm's answer \emph{``Basil Mahfouz Al-Kuwaiti''}.
 On the other hand, ROUGE-L score is 0.17 for both answers.
\label{app}

\begin{table}[h]
\renewcommand{\arraystretch}{1.2}
\begin{center}
\caption{Utility across datasets, LLM models and systems. Accuracy based on ROUGE-L recall score.}
\begin{tabular}{ r | c | c c}
\multicolumn{1}{c|}{\multirow{2}{*}{LLM}} & \multicolumn{1}{c|}{\multirow{2}{*}{System}}  & \multicolumn{2}{c}{Datasets}\\
\cline{3-4}
& & DMR & TOFU\\
\hline
\multicolumn{1}{c|}{\multirow{2}{*}{gpt-4.1-nano}}
& \memllm & 66.6\% & 12.0\% \\
& \dellm     & 64.2\% & 34.3\% \\
\hline
\multicolumn{1}{c|}{\multirow{2}{*}{gpt-5-nano}}
& \memllm & 60.5\% & 14.8\% \\
& \dellm     & 57.6\% & 24.4\% 
\end{tabular}
\label{tab:utility-r}
\end{center}
\end{table}

\begin{table}[h]
\begin{center}
\renewcommand{\arraystretch}{1.2}
\caption{Deletion rate across datasets, LLM models and systems. ROUGE-L recall scores reported.
A lower After score is preferred, as it indicates that the utility on the forget data has decreased following a deletion request for that data.}
\setlength{\tabcolsep}{3pt}
\begin{tabular}{ r | c | r r | r r }
 \multicolumn{1}{c|}{\multirow{3}{*}{LLM}} &
 \multicolumn{1}{c|}{\multirow{3}{*}{System}} &
 \multicolumn{4}{c}{Datasets} \\
\cline{3-6}
 & & \multicolumn{2}{c|}{TOFU} & \multicolumn{2}{c}{PF}\\
\cline{3-6} 
 & & Before & After & Before & After\\
\hline
\multicolumn{1}{c|}{\multirow{3}{*}{gpt-4.1-nano}}
& \nomemllm & 18.8 & 18.8 & 1.5 & 1.5 \\
& \memllm  & 13.0 & 13.0 & 75.8 & 65.3 \\
& \dellmbf    & 37.7 & 24.0 & 82.2 & 20.4 \\
\hline
\multicolumn{1}{c|}{\multirow{3}{*}{gpt-5-nano}}
& \nomemllm & 17.5 & 17.5 & 10.3 & 10.3 \\
& \memllm  & 15.5 & 15.0 & 83.2 & 77.5 \\
& \dellmbf    & 26.1 & 15.4 & 91.3 & 15.8 
\end{tabular}
\label{tab:delete-r}
\end{center}
\end{table}

\begin{table}[h]
\begin{center}
\renewcommand{\arraystretch}{1.2}
\caption{Impact of deletion of forget data on utility over retain dataset.  ROUGE-L recall scores reported.
Higher After score is preferred, as it indicates that the utility on the retain data has not been impacted significantly after the deletion request on forget data.}
\setlength{\tabcolsep}{3pt}
\begin{tabular}{ r | c | r r | r r }
 \multicolumn{1}{c|}{\multirow{3}{*}{LLM}} &
 \multicolumn{1}{c|}{\multirow{3}{*}{System}} &
 \multicolumn{4}{c}{Datasets} \\
\cline{3-6}
& & \multicolumn{2}{c|}{TOFU} & \multicolumn{2}{c}{PF}\\
\cline{3-6}
&& Before & After & Before & After \\
\hline
\multicolumn{1}{c|}{\multirow{2}{*}{gpt-4.1-nano}}
& \memllm & 11.1 & 11.1 & 70.3 & 66.9 \\
& \dellmbf   & 30.9 & 29.3 & 83.0 & 80.8 \\
\hline
\multicolumn{1}{c|}{\multirow{2}{*}{gpt-5-nano}}
& \memllm & 14.1 & 13.5 & 79.1 & 77.2 \\
& \dellmbf   & 22.6 & 23.8 & 87.4 & 86.3 
\end{tabular}
\label{tab:impact-r}
\end{center}
\end{table}

\subsection{A prompt to ChatGPT to generate PF dataset}
\label{app:prompts}

\begin{tcolorbox}[
    title={LLM Prompt},
    breakable,
    colback=gray!5,
    colframe=black!60,
    boxrule=0.5pt
]
\begin{Verbatim}[breaklines=true, breakanywhere=true, breaksymbol={}]
Generate a JSON object with 100 fictional persons using the following structure:

1. persona: An array containing:
   - 10 first-person statements describing the person, including useful personal facts such as age, occupation, location, family, interests, preferences, pets, and future plans.
   - Follow those statements with 5 natural questions/requests that the same person might ask an LLM. These questions should require the LLM to use information from the earlier statements to answer or perform the request properly.
   - Do not explicitly state which facts the LLM questions depend on.

2. qa: An array containing 4 questions that directly test facts stated in `persona`. Each object should contain:
   - "question": A question about the person.
   - "answer": The correct answer based on the description.
   - "fact": The statement in persona containing the answer.

Return valid JSON only.
\end{Verbatim}
\end{tcolorbox}

\end{document}